\documentclass[nohyperref]{article}

\usepackage{microtype}
\usepackage{graphicx}
\usepackage{subfigure}
\usepackage{booktabs} 
\usepackage{dirtytalk}
\usepackage{float}

\usepackage{hyperref}

\usepackage[accepted]{icml2023}

\usepackage{amsmath}
\usepackage{amssymb}
\usepackage{mathtools}
\usepackage{amsthm}

\usepackage{pifont}
\newcommand{\cmark}{\hspace*{\fill}{\color{green}\ding{51}}\hspace*{\fill}}%
\newcommand{\xmark}{\hspace*{\fill}{\color{red}\ding{55}}\hspace*{\fill}}%
\usepackage[capitalize,noabbrev]{cleveref}

\theoremstyle{plain}

\theoremstyle{definition}

\theoremstyle{remark}

\usepackage[textsize=tiny]{todonotes}

\icmltitlerunning{The Acquisition of Physical Knowledge in Generative Neural Networks}

\begin{document}

\twocolumn[
\icmltitle{The Acquisition of Physical Knowledge \\ in Generative Neural Networks}



\icmlsetsymbol{equal}{*}

\begin{icmlauthorlist}
    \icmlauthor{Luca M. Schulze Buschoff}{MPIBC}
    \icmlauthor{Eric Schulz}{MPIBC}
    \icmlauthor{Marcel Binz}{MPIBC}
\end{icmlauthorlist}

\icmlaffiliation{MPIBC}{MPRG Computational Principles of Intelligence, Max Planck Institute for Biological Cybernetics, Tübingen, Germany}
\icmlcorrespondingauthor{Luca Schulze Buschoff}{luca.schulze-buschoff@tuebingen.mpg.de}

\icmlkeywords{machine Learning, cognitive development, developmental psychology, intuitive physics, computational cognitive science, violation-of-expectation, generative neural networks}

\vskip 0.3in
]



\printAffiliationsAndNotice{} 

\begin{abstract}

As children grow older, they develop an intuitive understanding of the physical processes around them. Their physical understanding develops in stages, moving along developmental trajectories which have been mapped out extensively in previous empirical research. Here, we investigate how the learning trajectories of deep generative neural networks compare to children's developmental trajectories using physical understanding as a testbed. We outline an approach that allows us to examine two distinct hypotheses of human development -- stochastic optimization and complexity increase. We find that while our models are able to accurately predict a number of physical processes, their learning trajectories under both hypotheses do not follow the developmental trajectories of children. 

\end{abstract}

\begin{figure*}[!h]
        \centering
        \includegraphics[width=0.8\textwidth]{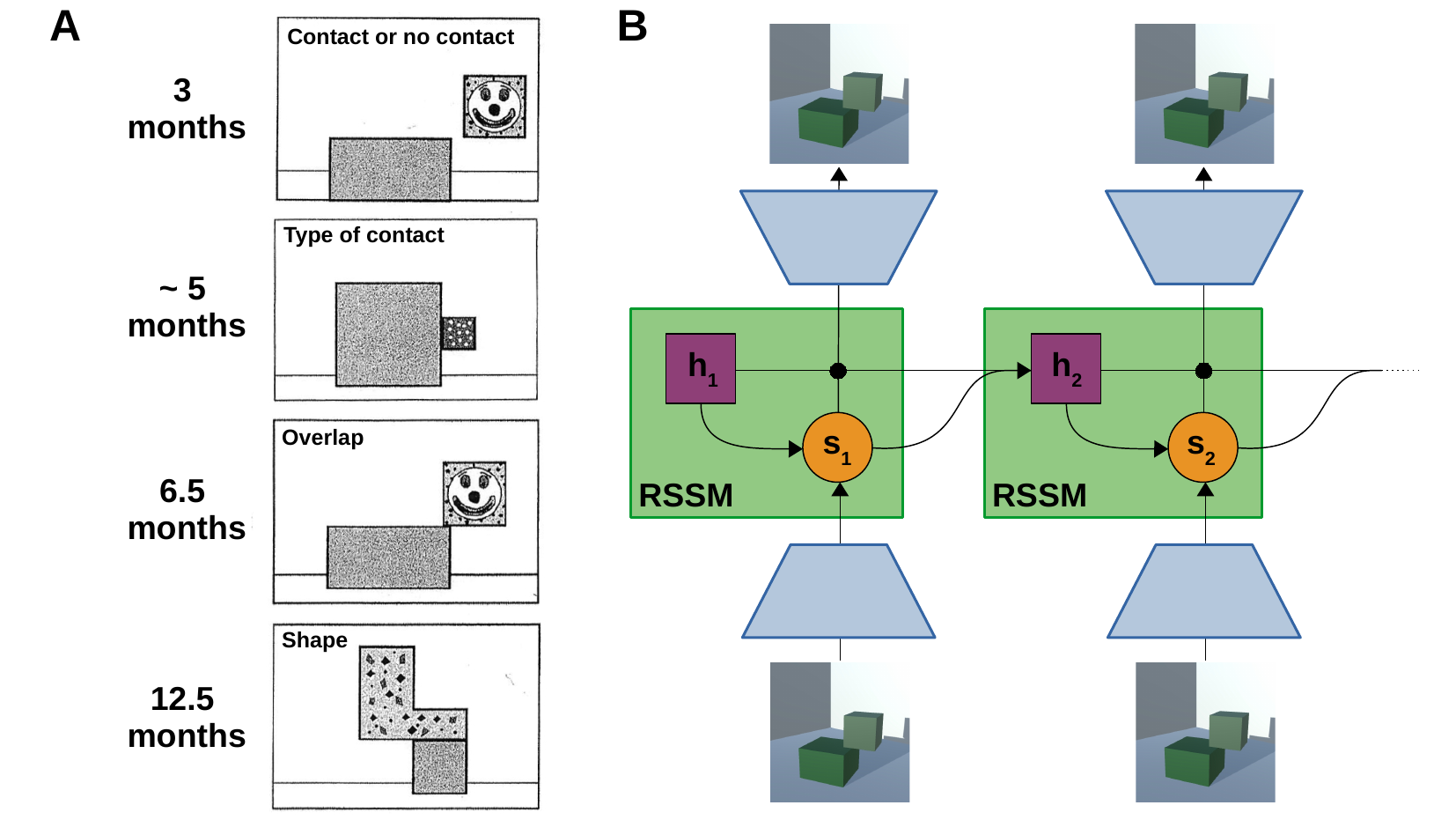}
        \caption{\textbf{A}: Human developmental trajectory for support events outlined by \citet{baillargeon1996infants}. The illustrations are taken from \citet{baillargeon1996infants} and they show the physical rules acquired at the respective ages. With $3$ months, infants decide based on a simple contact or no contact rule. According to this rule, a block configuration is considered stable if the blocks touch each other. At around $5$ months, infants understand that the type of contact matters. Now, only configurations with blocks stacked on top of each other are judged as stable. At $6.5$ months, they begin to also consider the overlap of the blocks. Finally, at $12.5$ months they are able to incorporate the block shapes into their judgement, relying not only on the amount of contact but also on how the mass is distributed for each block. \textbf{B}: Illustration of our generative video prediction model.}
        \label{fig:model_graph}
\end{figure*}
    
\section{Introduction}
\label{sec:introduction}

    More than 70 years ago, \citet{turing1950computing} famously suggested that \say{instead of trying to produce a programme to simulate the adult mind, why not rather try to produce one which simulates the child's? If this were then subjected to an appropriate course of education one would obtain the adult brain.} If we want to take Turing’s proposal seriously, we have to ask ourselves: how do children learn?

    Developmental psychologists have investigated children's learning in a number of different realms. One of the most well-studied is their acquisition of physical knowledge \citep{baillargeon1996infants, baillargeon2004infants, spelke2007core, lake2017building}. Here, prior empirical work provides us with a precise understanding of the stages that children undergo during their cognitive development (see Figure \ref{fig:model_graph}A for an example). It, therefore, serves as an ideal testbed for our investigation.

    In the present paper, we set out to formalize and test two distinct hypotheses of children's development. The first is the idea of \emph{development as stochastic optimization}, which argues that cognitive development results from some form of stochastic optimization procedure \citep{gopnik2017changes, ullman2020bayesian, giron2022developmental, wolff1987cognitive}. The second is the idea of \emph{development as complexity increase}, which instead stipulates that the knowledge structures involved in human reasoning become more complex over time \citep{baillargeon2002acquisition, binz2019emulating}.
    
    First, we show how both hypotheses can be instantiated in a $\beta$-variational autoencoder ($\beta$-VAE) framework. We then probe models with different degrees of complexity and optimization on physical reasoning tasks using violation-of-expectation (VOE) methods \citep{piloto2018probing, smith2019modeling}. Finally, we compare the learning trajectories of these artificial systems to the developmental trajectories of children. 
    
    We find that even fairly generic deep generative neural networks acquire many physical concepts. However, the order in which they acquire these concepts under both hypotheses does not align well with the acquisition order of children -- neither hypothesis fully captures the learning trajectories of children. Thus, we conclude that the investigated models do not acquire their knowledge in accordance with Turing's proposal.

    The remainder of this paper is organized as follows. Section 2 surveys previous literature on models of human-like physical knowledge and developmental trajectories. In Section 3, we illustrate how to instantiate the development as stochastic optimization and development as complexity increase hypotheses in the $\beta$-VAE framework. We then apply these models to different physical reasoning domains in Section 4. Section 5 concludes this report with a general discussion of our findings.

\section{Related work}
\label{sec:related}

\begin{table*}[ht]
     \caption{Table summarizing previous work attempting to build models with human-like physical intuitions and work attempting to model human developmental trajectories. Machine learning research has predominantly focused on reproducing adult-level performance, while computational cognitive science has relied heavily on low-dimensional and static stimuli. The present paper combines the best of both worlds.}
    \vspace{0.5cm}
    \centering
    \begin{tabular}{@{}p{3.5cm}p{1.1cm}p{1.3cm}p{1.3cm}p{1.65cm}p{1.65cm}p{1.6cm}p{1.82cm}}
        \toprule
         & support events & occlusion events & collision events & unsupervised learning & violation-of-expectation & sequential predictions & developmental trajectories\\
        \midrule 
        \multicolumn{7}{@{}l}{\textbf{Work attempting to build models with human-like physical intuitions:}} \\
        \midrule
        \citet{battaglia2013simulation} & \cmark & \xmark & \xmark & \xmark & \xmark & \xmark & \xmark  \\
        \citet{battaglia2016interaction} & \xmark & \xmark & \cmark & \xmark & \xmark & \cmark & \xmark \\
        \citet{smith2019modeling} & \xmark & \cmark  & \xmark & \cmark & \cmark  & \cmark & \xmark\\
        \citet{lerer2016learning} & \cmark & \xmark & \xmark & \xmark & \xmark & \xmark & \xmark\\
        \citet{zhang2016comparative} & \cmark & \xmark & \xmark & \xmark & \xmark & \xmark & \xmark\\
        \citet{piloto2018probing} & \xmark & \xmark & \xmark & \cmark & \cmark & \cmark & \xmark \\
        \citet{riochet2021intphys} & \xmark & \xmark  & \xmark  & \cmark & \cmark & \xmark  & \xmark  \\
        \citet{piloto2022learn} & \cmark & \cmark & \cmark & \cmark & \cmark & \cmark & \xmark \\
        \midrule
        \multicolumn{7}{@{}l}{\textbf{Work attempting to model human developmental trajectories:}} \\
        \midrule
        \citet{giron2022developmental} & \xmark & \xmark & \xmark & \xmark & \xmark & \xmark & \cmark \\
        \citet{averbeck2022pruning} & \xmark & \xmark & \xmark & \xmark & \xmark & \xmark & \cmark \\
        \citet{huber2022developmental} & \xmark & \xmark & \xmark & \xmark & \xmark & \xmark & \cmark \\
        \citet{binz2019emulating} & \cmark & \cmark & \xmark & \xmark & \xmark &\xmark & \cmark \\
        \midrule
        This work & \cmark & \cmark & \cmark & \cmark & \cmark & \cmark & \cmark\\
        \bottomrule
    \end{tabular}
    \label{tab:summary}
\end{table*}

    \subsection{Models with human-like physical knowledge}
    Building models with human-like physical knowledge has become an active research area in recent years (see Table \ref{tab:summary} for a summary). \citet{battaglia2013simulation} argued that human reasoning in complex natural scenes is driven by an intuitive physics engine that relies on probabilistic simulations to make inferences. Following this idea, they introduced interaction networks -- a model that performs simulations by combining an object-centric and a relation-centric component \citep{battaglia2016interaction}. In contrast to the initial approach that relied on a hard-coded physics engine, interaction networks are learnable engines, allowing them to generalize to novel systems with different configurations of objects and relations. In a similar vein, \citet{smith2019modeling} combined a perception module that infers physical object representations from raw images with a reasoning module that predicts future object states conditioned on the object representations. They found that this model matched human performance in a number of scenarios. \citet{lerer2016learning} trained large convolutional neural networks to predict the stability of wooden block towers as well as the trajectories of falling blocks. They showed that the performance of such networks exceeds that of human subjects on synthetic data. \citet{zhang2016comparative} compared the intuitive physics engine of \citet{battaglia2013simulation} to the convolutional neural network of \citet{lerer2016learning}. They found that while convolutional networks are able to achieve superhuman accuracy in judging the stability of block towers, their physical understanding is dissimilar to that of humans.

    \subsection{The violation-of-expectation method}
    How physical knowledge of artificial systems should be evaluated has also received attention. Taking inspiration from developmental psychology, \citet{piloto2018probing} proposed to use the VOE method to probe the knowledge of neural networks \citep{baillargeon1996infants}. In particular, they measured the surprise of a network after observing physically implausible sequences. Their work was among the first to demonstrate that the VOE method can elucidate black-box models' inference mechanisms. Moreover, recent intuitive physics benchmarks have also been inspired by work in developmental psychology. \citet{riochet2021intphys} presented an \say{evaluation benchmark which diagnoses how much a given system understands about physics by testing whether it can tell apart well-matched videos of possible versus impossible events constructed with a game engine.} Likewise, \citet{weihs2022benchmarking} proposed a benchmark testing for knowledge about continuity, solidity, and gravity using videos filmed in infant-cognition labs and robotic simulation environments. Finally, \citet{piloto2022learn} also introduced a data set for evaluating intuitive physics in neural networks using the VOE method and use this data set to probe the physical knowledge of a deep learning model equipped with object-centric representations. 
    
    \subsection{Modeling human development}   
    Even though developmental psychology has inspired how to evaluate physical knowledge in neural networks, the emphasis of prior machine learning research has always been on reproducing adult-level performance. In contrast, computational cognitive scientists also strive to build artificial learning systems that capture the developmental trajectories of children. Perhaps most closely related to our work is the approach of \citet{binz2019emulating} who compared trajectories of Bayesian neural networks that had access to different amounts of data to human developmental trajectories. They investigated both occlusion and support events and found that the acquisition order of concepts in their model aligned with that of children. However, in contrast to their work, which uses an oracle to provide a supervision signal about block stability and visibility, our approach solely relies on an unsupervised training objective.

    If we look beyond the realm of intuitive physics, we can find other works that have attempted to model the process of human development. \citet{huber2022developmental} investigated the emergence of object recognition in children. They showed that four- to six-year-olds are already more robust to image distortions compared to deep neural networks trained on ImageNet. Furthermore, children predominantly relied on shape instead of texture for object detection, making them more similar to adults than deep neural networks \citep{geirhos2018imagenet}. \citet{averbeck2022pruning} pruned recurrent neural networks by removing weak synapses. They found that pruned networks were more resistant to distractions in a working memory task and made optimal choices more frequently in a reinforcement learning setting. These results were consistent with developmental improvements during adolescence, where performance on cognitive operations improves as excitatory synapses in the cortex are pruned. Finally, \citet{giron2022developmental} examined a theory of development as stochastic optimization. In particular, they combined this idea with a model of human decision-making in multi-armed bandit problems and demonstrated that development resembles a stochastic optimization process in the parameter space of this model. In contrast to these earlier models of development, our setup uses high-dimensional visual stimuli (i.e., video sequences) and solely relies on an unsupervised training objective. It, therefore, more closely resembles the actual learning processes of children in the real world. 

\section{Methods}  \label{sec:model}
    
        In the following, we discuss how the \emph{development as stochastic optimization} and \emph{development as complexity increase} hypotheses can be instantiated in the $\beta$-VAE framework\footnote{We do not propose that these exact mechanisms are implemented by the brain. Rather, we propose these mechanisms as "as if" models: while the mechanisms behind biological development are likely not exactly analogous, they have similar characteristics.}. For the development as stochastic optimization hypothesis, we train a generative video prediction model using gradient descent. To obtain a learning trajectory of this model, we evaluate snapshots of the model in every epoch. For the development as complexity increase hypothesis, we train models of different complexities by making use of the $\beta$-VAE framework \citep{higgins2016beta}. Doing so enforces a bottleneck on the representational capacity of the hidden representations \citep{sims2016rate, bates2020efficient}, which can be interpreted as a particular form of computational complexity (we will discuss potential alternatives in our general discussion). Learning trajectories for this hypothesis are obtained by increasing the model's representational capacity, i.e., by moving from higher to lower $\beta$-values within fully converged models.

        \subsection{Model architecture and objective}
        We use the recurrent state space model (RSSM) \citep{hafner2019learning, saxena2021clockwork} -- which can be seen as a sequential version of a VAE -- as an exemplary model for our analysis. We selected this particular model for two reasons. First, generative models are, in general, a good starting point as they have previously been used to model intuitive physics \citep{piloto2018probing, piloto2022learn, riochet2021intphys} (see also \citep{duan2022survey} for an overview). Furthermore, we decided on a VAE-based model since such models have been shown to capture many aspects of human cognition \citep{malloy2022modeling, nagy2020optimal, bates2020efficient}, thereby making them a reasonable candidate hypothesis for studying human development. The model maintains a latent state at each time step, which is comprised of a deterministic component $h_t$ and a stochastic component $s_t$ (see Figure \ref{fig:model_graph}B). These components depend on the previous time steps through a function $f(h_{t-1}, s_{t-1})$, which is implemented as a gated recurrent neural network. We train our models by optimizing the following objective:
        
        \begin{align} \label{eq:rssm_loss}
            -\sum_{t=1}^T & ~\mathbb{E}_{q(s_t\mid o_{\leq t})}[\textrm{ln } p(o_t\mid s_t)] + \\
            \beta & ~\mathbb{ E}_{q(s_{t-1}\mid o_{\leq t-1})}\big[\textrm{KL} \big( q(s_t \mid o_{\leq t}) \mid \mid p(s_t\mid s_{t-1}) \big) \big]    \nonumber
        \end{align} 

        where $o_{\leq t} = o_1, o_2, \ldots, o_t$ is a sequence of rendered images obtained from a 3D physics engine.

        For all models, the size of the stochastic hidden dimension $s_t$ was kept at $20$, while the size of the deterministic hidden dimension $h_t$ was set to $200$, as in previous implementations of the RSSM \citep{hafner2019learning, saxena2021clockwork}. We furthermore adopted the image encoder and decoder architectures described by \citet{dittadi2020transfer}. We refer the reader to Appendix \ref{app:model} for further details about the model architecture and training procedure. 
        
        We can use the RSSM to generate either open- or closed-loop predictions. For open-loop predictions, the model processes a number of initial observations to infer an approximate posterior $q(s_{t-1}\mid o_{\leq t-1})$, followed by decoding subsequent latent representations sampled from the prior $p(s_t\mid s_{t-1})$. For closed-loop reconstructions, the decoder is instead continuously given representations sampled from the posterior which is updated at every time step using the previously observed frame. We generally report results obtained via open-loop predictions unless stated otherwise.

    \subsection{Measuring surprise}
    \label{sec:measuring_surprise}
    
        To assess whether a model has learned a specific physical rule, we make use of the VOE paradigm \citep{piloto2018probing, piloto2022learn}. For this, the model is presented with two video sequences: a \emph{violated} sequence, which constitutes a violation according to the rule, and an \emph{expected} sequence, which is consistent with the rule. If the model has successfully learned a specific rule, it should show a larger degree of surprise for the violated compared to the expected sequence. 
        
        Following \citet{piloto2022learn} and \citet{smith2019modeling}, we measure the model's surprise using the decoder’s negative log-likelihood (NLL) of observations (i.e., the first term of Equation \ref{eq:rssm_loss})\footnote{Note that this definition differs from the common usage of surprise in the context of information theory, where it is defined to be $-\log p (o \leq _t)$.}. More specifically, for each sequence, we determine if the NLL is larger for the violated compared to the expected sequence for the majority of the frames. We then take the mean over all sequences for a specific condition in order to check whether the reconstructions of the model better match the expected or the violated sequences. This approach is inspired by developmental psychology and allows us to measure a model's surprise similar to how developmental psychologists measure surprise in children (see Appendix \ref{app:surprise} for a discussion on different measures of surprise).

\section{Results}
\label{sec:results}

    \begin{figure*}[!t]
        \centering
        \includegraphics[width=0.9\textwidth]{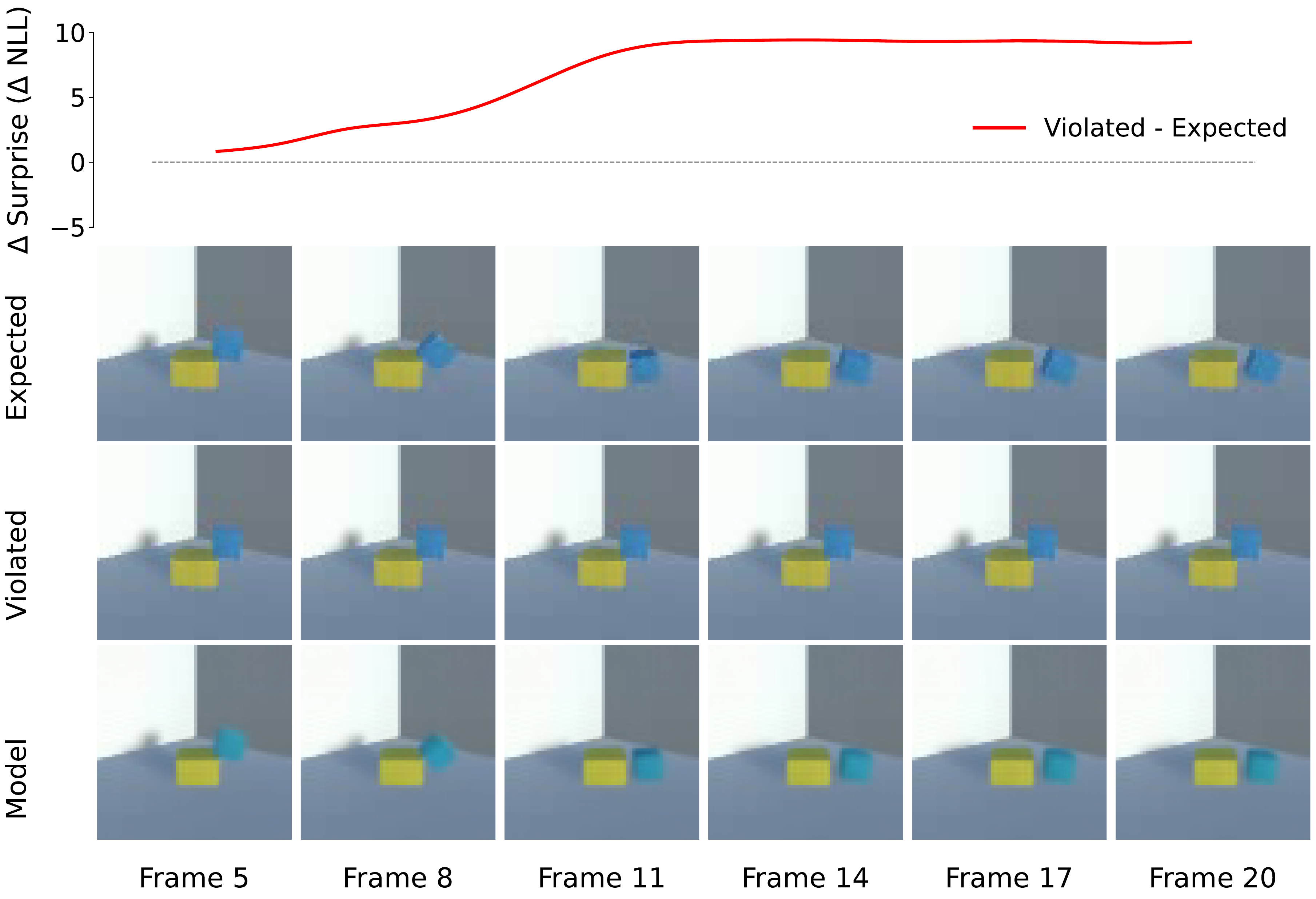}
        \caption{The first row shows the difference in surprise between the violated and expected test sequences given by the fully trained model with $\beta$ = 1 for the overlap condition in the support event data-set. The surprise curve is smoothed using cubic spline interpolation. The second row shows the expected test sequence. The third row shows the violated test sequence. The last row shows the open-loop reconstruction from the model given the first two frames.}
        \label{fig:support_lloverframes}
    \end{figure*}
    
    We evaluated our models on three distinct physical processes. For each of these processes, we generated training data sets inspired by experiments from developmental psychology using the Unity game engine \citep{unity}. We randomly varied a number of properties to ensure sufficient variability in the training data. We also generated test data sets that -- following the VOE paradigm -- contain pairs of violated and an expected sequences for each of the conditions in the respective event types (see Appendix \ref{app:datasets} for a detailed description of the data generation process and a visualization of the employed test sequences).
   
    \begin{figure*}[!t]
            \centering
            \includegraphics[width=1.0\textwidth]{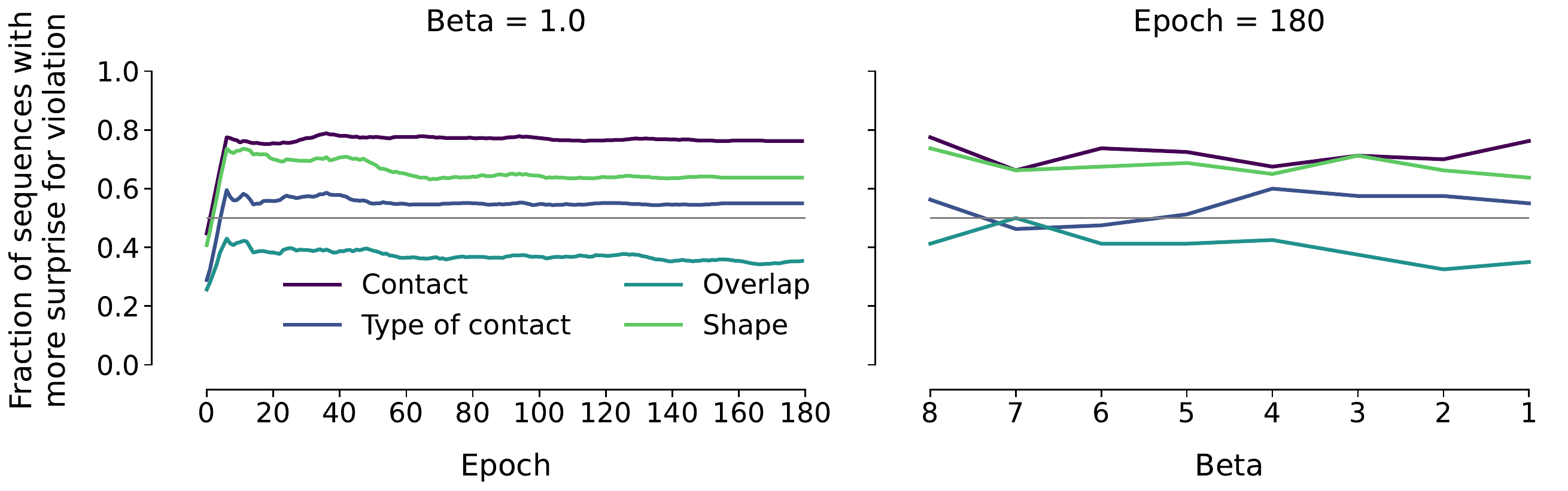}
            \vspace{-0.5cm}
            \caption{The plot on the left shows the percentage of sequences for which the surprise for the violated sequence exceeds that of the expected sequence for the model with $\beta = 1$ at every epoch and for each condition of the support event data set. The lines are smoothed with a uniform kernel of size 10. The plot on the right shows the same metric for fully trained models with different $\beta$.}
            \label{fig:support_hypotheses}
    \end{figure*}
        
    \subsection{Support events}

        We began our investigations by looking at support events, which consist of block configurations such as the ones shown in Figure \ref{fig:model_graph}A. Similar tasks have been studied extensively in both the machine learning and developmental psychology community, and they, therefore, serve as an ideal starting point for our analyses. Each scene in our data set contains two randomly configured blocks in a gray room.\footnote{Note that it would certainly be possible to consider more complex configurations (e.g., by increasing the number of blocks), but we deliberately made this design choice to match the experimental paradigms used in developmental psychology.}
        
        \citet{baillargeon1996infants} has shown that, as infants grow older, they make use of increasingly complex rules to decide whether a given block configuration is stable or not (also see \citet{baillargeon2002acquisition, baillargeon2004infants}). With $3$ months, infants decide based on a simple contact or no contact rule. According to this rule, a block configuration is considered stable if the blocks touch each other. At around $5$ months, infants understand that the type of contact matters. Now, only configurations with blocks stacked on top of each other are judged as stable. At $6.5$ months, they begin to also consider the overlap of the blocks. Finally, at $12.5$ months they are able to incorporate the block shapes into their judgement, relying not only on the amount of contact but also on how the mass is distributed for each block. 
        
        For each of the four rules for support events, we constructed pairs of violated and expected test sequences with identical first frames. For example, according to the overlap rule, a block configuration should only be stable if the blocks are stacked on top of each other with enough overlap. A test sequence pair for this rule shows two blocks that only slightly overlap. The expected test sequence, which is consistent with the rule, shows the top block falling. In contrast -- and in violation of real physics -- the violated test sequence shows the same block configuration that however appears stable (Figure \ref{fig:support_lloverframes} shows an example pair for this rule, together with the predictions of our model and its corresponding surprise values).
        
        We first verified that our model is able to predict a given scene accurately into the future. For this purpose, we plotted the open-loop predictions given by the fully trained model with $\beta=1$. Figure \ref{fig:support_lloverframes} shows an examplary result for the overlap condition. We see that the predictions of the fully trained model closely match the expected sequence. Furthermore, we see that high surprise values for the violated sequence coincide with differences to the expected sequence -- the model is surprised when it observes parts of a video sequence that diverge from real physics. Appendix \ref{app:support_recons} shows further examples for open- and closed-loop predictions, while Appendix \ref{app:recon_errors} contains a visualization of prediction errors.
                
        Figure \ref{fig:support_hypotheses} illustrates how knowledge about physical rules develops over time for the two earlier outlined hypotheses. On the left, the percentage of sequences for which the surprise for the violated sequence exceeds that of the expected sequence is plotted for each of the four conditions over the course of training for the model with $\beta$ = 1. Here, the model becomes increasingly optimized over the epochs, thereby implementing the development as stochastic optimization hypothesis. It is evident that the model is able to learn three of the four conditions as it shows more surprise for the violated than the expected sequences for the majority of the cases. However, it learns the conditions at roughly the same rate which does not match the developmental trajectories of children. While it settles at different levels for the conditions, the order of these conditions also does not match the acquisition order of children: the shape condition, for instance, shows the second highest percentage while it is the last rule that children acquire. 
        
        On the right, the percentage of sequences for which the surprise for the violated sequence exceeds that of the matching expected sequence is plotted for each of the four conditions for fully trained models with different $\beta$-values. This relates to the development as complexity increase hypothesis since the representational capacity of the model increases as $\beta$ decreases. The order in which increasingly complex models learn the different conditions again does not resemble the developmental trajectories of children: the model with $\beta = 8$ performs very similarly to the model with $\beta = 1$. To summarize, for support events, neither hypotheses yields learning trajectories that resemble the developmental trajectories of children (see also Appendix \ref{app:closed} for a closed-loop counterpart to Figure \ref{fig:support_hypotheses}). 
        

    \begin{figure*}[!t]
        \centering
        \includegraphics[width=1.0\textwidth]{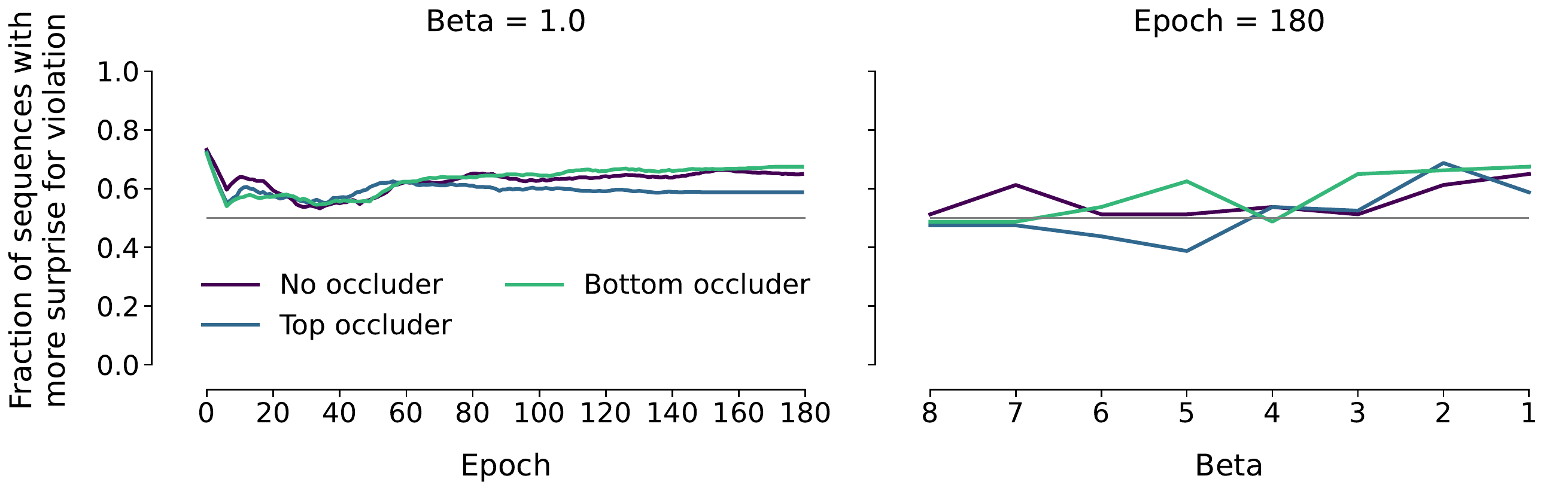}
        \vspace{-0.5cm}
        \caption{The plot on the left shows the percentage of sequences for which the surprise for the violated sequence exceeds that of the expected sequence for the model with $\beta = 1$ at every epoch and for each condition of the occlusion event data set. The lines are smoothed with a uniform kernel of size 10. The plot on the right shows the same metric for fully trained models with different $\beta$.}
        \label{fig:occlusion_hypotheses}
    \end{figure*}
    
    \subsection{Occlusion events}
    
        Next, we wanted to test whether the results obtained in the last section hold across domains. Thus, we extended our analyses to occlusion events, which display a moving object passing behind two vertical columns. The two columns, together with an optional horizontal connection at the top or bottom, form an occluder which may hide the moving object. Like in the preceding section, we created a randomized training data set alongside several test sequences that violate physical principles (see Appendix \ref{app:datasets} for examples and further details about the data generation process).

        \citet{baillargeon1996infants} reported that, in this setting, (1) infants form a simple behind/not-behind distinction by 2.5 months. Hereby they assume that the object will not re-appear in the gap between the columns that are connected at the top or bottom.  By 3 months, (2) infants expect objects to re-appear when the columns are connected at the top but fail to do so if the columns are connected at the bottom. Finally, at 3.5 months, (3) they also expect objects to appear behind screens that are connected at the bottom, given that the object is taller than the connecting part.

        Figure \ref{fig:occlusion_hypotheses} visualizes our modeling results by showing the percentage of sequences for which the surprise for the violated sequence exceeds that of the expected sequence for each condition. First, we can observe that the fully trained model with $\beta=1$ is surprised when presented with any of the test sequences that violate physical principles, indicating that it understood all of the three aforementioned occlusion settings. For the left side of the plot, which depicts the development as stochastic optimization hypothesis, we see that the number of sequences for which the surprise for the violated sequence exceeds that of the matching expected sequence increases at the same rate for all three conditions, meaning that the model learns the three concepts at approximately the same time. Thus, for occlusion events, the stochastic optimization hypothesis again does not yield a learning trajectory that matches that of children.
    
        The right side of Figure \ref{fig:occlusion_hypotheses} relates to the development as complexity increase hypothesis. Here, we see that the the percentage of sequences for which the surprise for the violated sequence exceeds that of the matching expected sequence remains relatively stable as the complexity of the model increases. This again does does not lend support to the complexity increase hypothesis.
        
    \subsection{Collision events}

    \begin{figure*}[!t]
        \centering
        \includegraphics[width=1.0\textwidth]{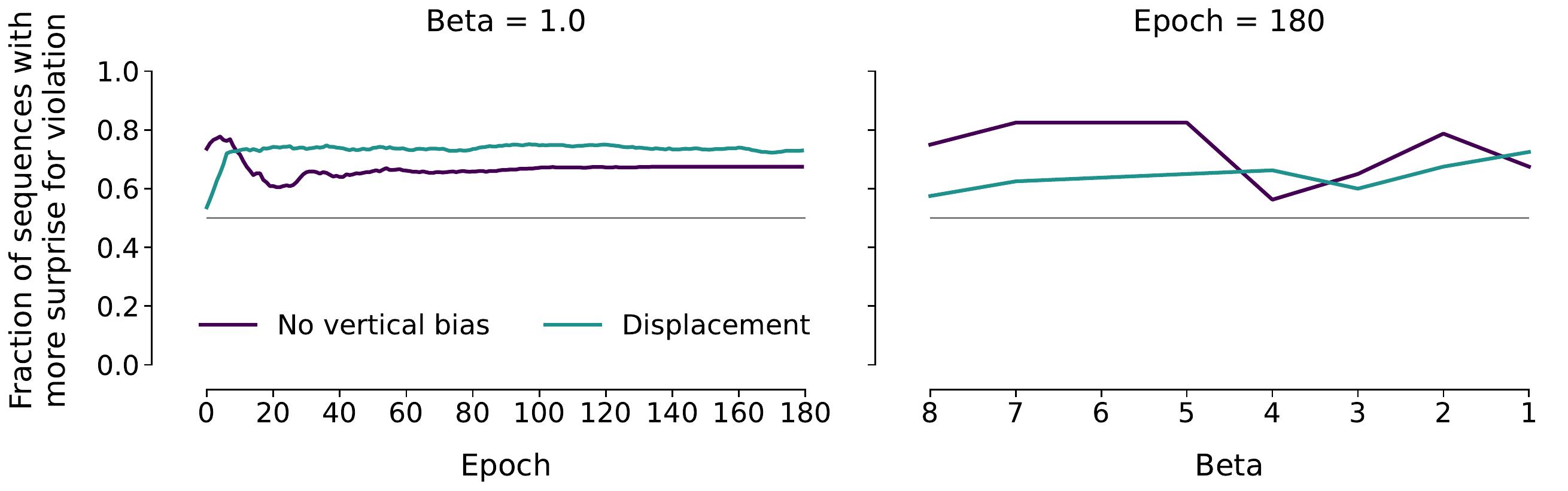}
        \vspace{-0.5cm}
        \caption{The plot on the left shows the percentage of sequences for which the surprise for the violated sequence exceeds that of the expected sequence for the model with $\beta = 1$ at every epoch and for each condition of the collision event data set. The lines are smoothed with a uniform kernel of size 10. The plot on the right shows the same metric for fully trained models with different $\beta$.}
        \label{fig:collision_hypotheses}
    \end{figure*}
    
        The last physical process that we investigated were collision events. Here, each scene shows an object rolling down a hill and colliding with a stationary object. To train our models, we again created a randomized training data set of such scenarios together with several test sequences that violate physical principles (see Appendix \ref{app:datasets} for examples and further details about the data generation process).
    
        \citet{baillargeon1996infants} provides two insights when it comes to collision events: (1) at first, infants expect any stationary object that collides with a moving object to be displaced by the same amount. However, as they grow older, they are able to take the relative sizes of the two objects into account and understand that the larger the size of the moving object compared to the stationary object, the larger the displacement of the stationary object. (2) Furthermore, at around 8 months, infants become subject to a \emph{vertical bias}, meaning that they judge stationary objects as immovable if they have a salient vertical dimension \citep{wang2003should, wang2004image}.

        Figure \ref{fig:collision_hypotheses} shows the learning trajectories of the two hypotheses, again by plotting how the percentage of sequences for which the surprise for the violated sequence exceeds that of the expected sequence develops over training epochs and complexity. Empirical research suggests that children first expect a size-independent displacement for all objects. To test whether our models exhibit such a characteristic, we compared their predictions for violated sequences with size-independent displacement (thereby violating physical principles) and expected sequences with a size-dependent displacement (working according to normal physics). While children initially show higher surprise when observing the expected sequences, our models offer a very different picture: at no point do they show more surprise for the expected compared to the violated sequences, as apparent by the plot on the left side of Figure \ref{fig:collision_hypotheses}.

        To test for the vertical bias, we constructed violated sequences where vertical objects do not move upon a collision and expected sequences where they do move according to normal physics. When presented with such sequence pairs, children show more surprise for the expected compared to the violated sequences at some point during their development. However, our models do not exhibit this characteristic at any point in time. Throughout training, they are more surprised by the violated compared to the expected sequences. Likewise, we also do not observe this effect when manipulating the representational capacity of our models as shown on the right side of Figure \ref{fig:collision_hypotheses}. For collision events, we therefore again find that neither the development as stochastic optimization nor the development as complexity increase hypothesis yield learning trajectories that resemble the developmental trajectories of children.

\section{Discussion}
\label{sec:discussion}

    We have compared the learning trajectories of an artificial system to the developmental trajectories of children for three physical processes. For this purpose, we outlined an approach that allowed us to investigate two distinct hypotheses of human development: stochastic optimization and complexity increase. 

    We found that the learning trajectories under both hypotheses do not follow children's developmental trajectories. For all three event types, we found differences to human learning. For support and occlusion events, the predictions of our models improve at roughly the same rate for all conditions, which indicates that our models do not move along separate stages. For collision events, our models crucially exhibit none of the biases that appear in children. We argue that this is to be expected. The vertical bias, for example, is likely a product of their self-directed movement in the world: as children begin to move around, the majority of vertical objects they encounter, such as walls or furniture, are immovable \citep{baillargeon1996infants}. In contrast to this, our models do not have access to such experiences and are therefore not incentivized to show this bias.
   
    While previous work on modeling cognitive development \citep{binz2019emulating, giron2022developmental} focused on tasks with low-dimensional and static stimuli, our approach employs high-dimensional visual stimuli (e.g., video sequences) and solely relies on an unsupervised training objective. It, therefore, more closely mirrors the actual learning processes of children in the real world. We furthermore extend previous research on building models with human-like physical intuitions by not focusing on adult-level performance but instead investigating developmental trajectories (see again Table \ref{tab:summary} for a comparison to previous research). 
   
    To showcase how our approach functions as a general framework for testing the learning trajectories of artificial systems, we used a fairly generic generative model. It would be interesting to evaluate the two hypotheses for other model classes, such as generative adversarial networks \citep{goodfellow2020generative} or diffusion models \citep{sohl2015deep} such as a denoising diffusion probabilistic model \citep{ho2020denoising}. Furthermore, it has been argued in previous work that object-centric representations are crucial for a proper physical understanding of more complex scenes \citep{piloto2022learn}. However, our models did not feature explicit object-centric representations and were still able to predict a number of physical processes. Thus, future work should aim for a systematic comparison of models with and without explicit object-centric representations. Additionally, future work should ideally incorporate model-based analyses investigating how changes in the hidden state are related to models' inability to capture the developmental trajectories of children (we have performed a rudimentary first analysis towards this end, see section \ref{app:tsne} in the Appendix).

    We used very simple data sets to determine the viability of our approach. Evidently, children do not learn by looking at a large number of stylized sequences. Instead, they observe the real world and generalize their acquired knowledge to a given experimental setting. To capture this process, future research should ideally train models in a similar way. This could, for example, be accomplished by utilizing the SAYCam data set \citep{sullivan2021saycam}. It contains a large number of longitudinal video recordings from infants’ perspectives. We believe that using this data set, it might be possible for an artificial model to acquire a vertical bias. It additionally includes time stamps indicating when a child has encountered a particular scene, which could be used to investigate how the nature of the training data influences development. It is also possible that children's pursuit of goals has an influence on their learning trajectories. Maybe this could also be investigated using the SayCAM dataset, however it is possible that collecting new data is required for answering this question.

    Finally, the complexity constraint we impose is a constraint on the size of the latent representations of the model. However, it is entirely possible that other parts of children’s physical models change in complexity throughout their development. For example, \citet{binz2019emulating} implement complexity increase through varying the complexity of model weights instead of the complexity of latent representations. In contrast to our work, they found that the acquisition order of concepts in their model aligned with that of children for support and occlusion events. Besides relying on a different complexity measure, their study diverged from ours in two additional ways: they used a much simpler set of stimuli (2D instead of 3D environments) and they relied on much less sophisticated models (plain feedforward networks instead of our sequential RSSM). We think an important question that needs to be addressed in future work is which of these design choices causes the divergence in results.

    What do we make of our results on the whole? On the one hand, they demonstrate that it is possible to use tools developed in psychology to elucidate the inner workings of deep learning models \citep{ritter2017cognitive, binz2022using}. From this perspective, our work highlights yet another mismatch between human learning and learning in artificial neural networks \citep{flesch2018comparing, dekker2022curriculum}. On the other hand, our results also indicate that current modeling approaches are quite far away from implementing Turing’s proposal for obtaining a programme that simulates the adult mind. If we want to keep following this direction, we have to therefore ask ourselves what is needed to build models that acquire their knowledge in human-like ways. Towards this end, it is possible that the training data plays an important role, as suggested by some of our results. However, it might be equally plausible that we need to develop new model architectures and come up with more sophisticated ways to train them.

\section*{Acknowledgements}
This work was funded by the Max Planck Society and the Volkswagen Foundation under grant number VW98569.

\clearpage
\bibliography{example_paper}
\bibliographystyle{icml2023}


\newpage
\appendix
\onecolumn

\section{Model implementation and training details}
\label{app:model}
    The models were implemented in PyTorch \citep{paszke2019pytorch}. For all models, the size of the stochastic hidden dimension $s_t$ was kept at 20, while the size of the deterministic hidden dimension $h_t$ was set to 200, as in previous implementations of the RSSM \citep{hafner2019learning, saxena2021clockwork}. 
    
    We used the encoder and decoder from \citet{dittadi2020transfer}. The encoder consists of 3 blocks. The first block consists of a convolutional layer with a kernel of size 5 and a stride of 2 and a padding of 2, followed by a leaky ReLU activation function, followed by 2 residual blocks. The second block consists of a convolutional layer with a kernel of size 1 and a stride of 1 and no padding, followed by average pooling with a kernel of size 2, followed by 2 blocks residual blocks. The third block consists of average pooling with a kernel of size 2, followed by 2 residual blocks. The fourth block consists of a convolutional layer with a kernel of size 1 and a stride of 1 and no padding, followed by average pooling with a kernel of size 2, followed by 2 residual blocks. The fifth block consists of average pooling with a kernel of size 2, followed by 2 residual blocks.
    
    The decoder consists of 5 blocks. The first block consists of 2 residual blocks, followed by upsampling with a scale factor of 2. The second block consists of 2 residual blocks, followed by a deconvolutional layer with a kernel size of 1 and a stride of 1, followed by upsampling with a scale factor of 2. The third block again consists of 2 residual blocks, followed by upsampling with a scale factor of 2. The fourth block consists of 2 residual blocks, followed by a deconvolutional layer with a kernel size of 1 and a stride of 1, followed by upsampling with a scale factor of 2. The fifth block consists of 2 residual blocks, followed by upsampling with a scale factor of 2, a leaky ReLU activation funktion, followed by a deconvolutional layer with a kernel size of 5 and a stride of 1 and a padding of 2.
    
    The models were trained for 180 epochs using a batch size of 32. The loss function was optimized using the Adam optimiser with a learning rate of 0.001 \citep{kingma2014adam}, which was divided by 10 every 50 epochs. The models were trained on a NVIDIA Quadro RTX 5000 for roughly 7 days. Our implementation of the RSSM borrows from a previous implementation on \href{https://github.com/cross32768/PlaNet_PyTorch}{GitHub}. The complete code for this project, including our model implementation, is available upon request.

\section{Different measures of surprise}
\label{app:surprise}

\begin{figure*}[!h]
        \centering
        \includegraphics[width=1.0\textwidth]{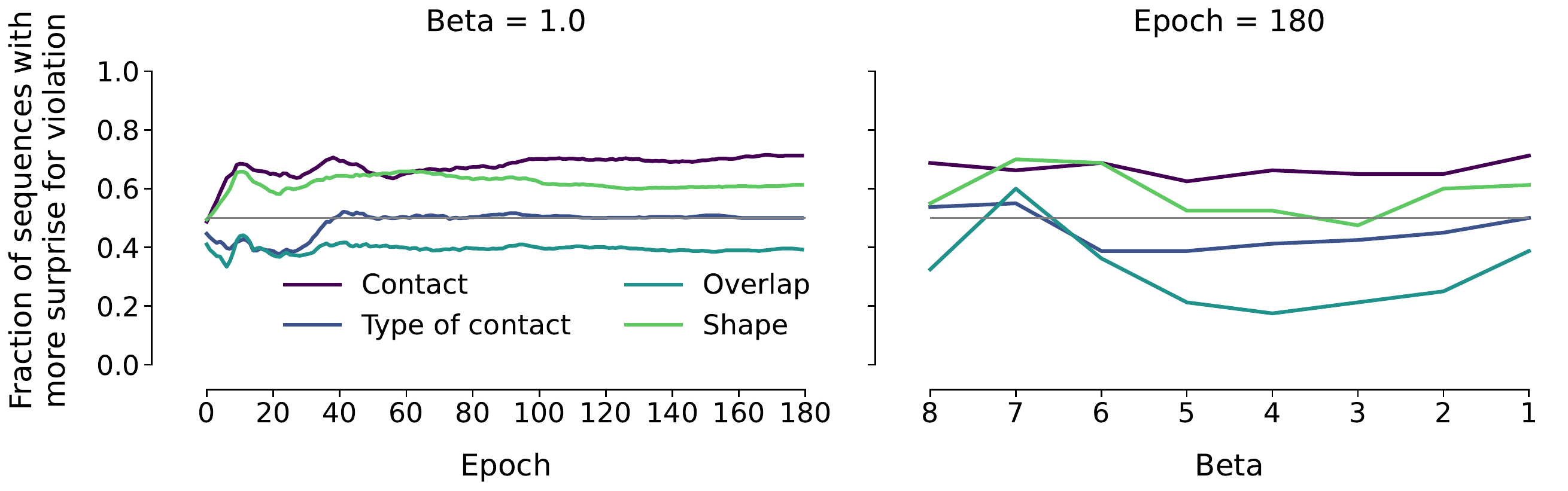}
        \vspace{-0.5cm}
        \caption{Replication of Figure \ref{fig:support_hypotheses} using the KL-based surprise measure. The plot on the left shows the percentage of sequences for which the surprise for the violated sequence exceeds that of the expected sequence for the model with $\beta = 1$ at every epoch and for each condition of the support event data set separately. The lines are smoothed with a uniform kernel of size 10. The plot on the right shows the same metric for fully trained models with different $\beta$.}
        \label{fig:support_hypotheses_kl}
    \end{figure*}

\begin{figure*}[!h]
        \centering
        \includegraphics[width=0.9\textwidth]{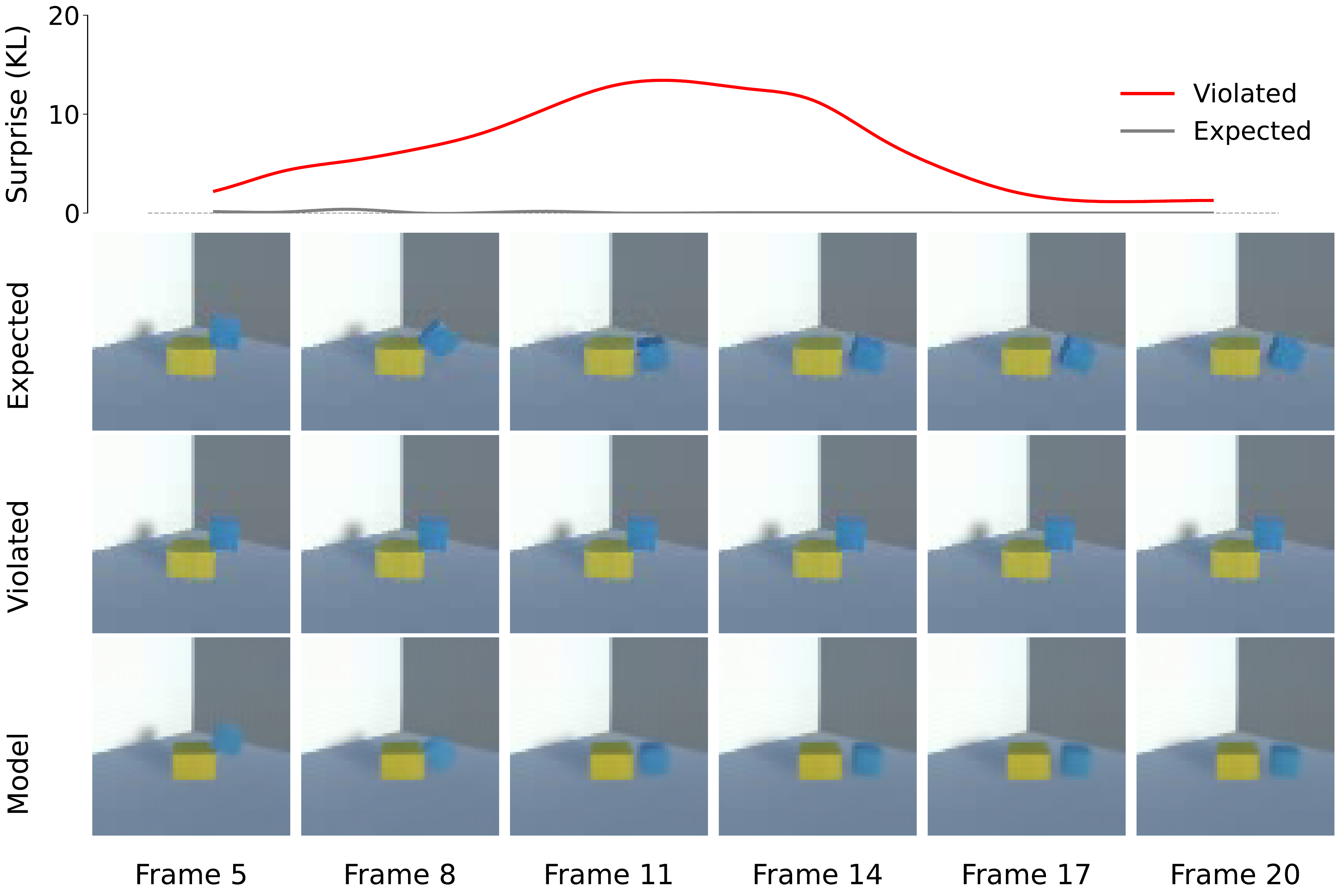}
        \caption{Replication of Figure \ref{fig:support_lloverframes} using the KL-based surprise measure. The first row shows the surprise given by the fully trained model with $\beta$ = 1 for the displayed expected and violated sequence. The surprise curves are smoothed using cubic spline interpolation. The second row shows the expected test sequence. The third row shows the violated test sequence. The last row shows the open-loop reconstruction from the model. The first four frames were removed for this plot.}
        \label{fig:support_kloverframes}
    \end{figure*}

    \citet{smith2019modeling} measure surprise as the maximum of the negative log-likelihood of observations under the model. Likewise, \citet{piloto2022learn} use the sum of the squared error, which given a Gaussian distribution with a standard deviation of one also equals the negative log-likelihood of observations under the model, up to a constant. We report the same measure in the main paper. However, \citet{piloto2018probing} propose another surprise measure: the KL-divergence between the prior and posterior over the latent representation \citep{baldi2010bits}. We confirmed our results using this measure and found only slight differences between the two measures (see Figures \ref{fig:support_hypotheses_kl} and \ref{fig:support_kloverframes} as an example).

    \vspace{-0.25cm}
    
\section{Training and test data sets}
\label{app:datasets}

    Each of the three event types is split into a training data set and a test data set. The training data set features 100.000 video sequences which each consist of 20 frames with a size of [64, 64, 3]. It was randomly split into 99.000 training sequences and 1000 validation sequences. The test data sets feature 80 pairs of expected and violated video sequences for each of the individual conditions in the respective event types. For the support event types, this results in a test data set with 640 video sequences. For the occlusion event types, the test data set consists of 480 video sequences. Finally, the test data set for the collision event types features 320 video sequences. The video sequences again consist of 20 frames with a size of [64, 64, 3].

\begin{figure*}[!h]
            \centering
            \includegraphics[width=0.8\textwidth]{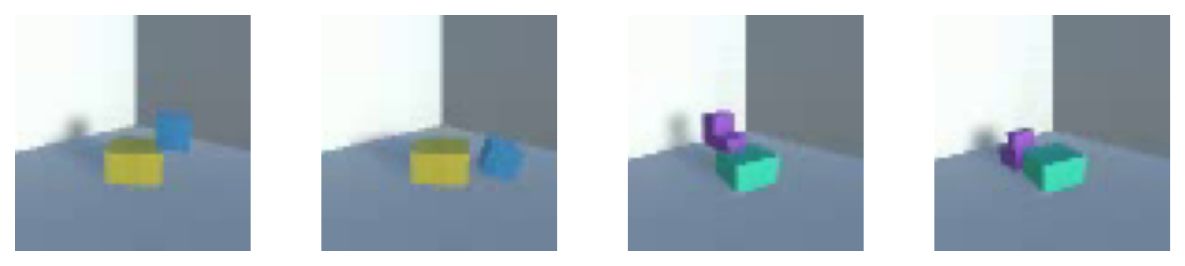}
            \vspace{-0.25cm}
            \caption{The last frame for example sequences from the support event test data set. From left to right and top to bottom, the conditions are \emph{contact or no contact}, \emph{type of contact}, \emph{overlap}, and \emph{shape} with each a violated sequence left and an expected sequence right.}
            \label{fig:support_test_example}
        \end{figure*}

    For the support events, the following variations were performed in order to ensure sufficient variability in the data sets: lower block size, lower block color, upper block color, lower block rotation, upper block rotation, upper block position (offset), and camera angle (see Figure \ref{fig:support_test_example} for more exemplary test sequences). For the training data set the shape of the upper block was also varied: half of the trials featured a cube as an upper block, while the other half featured an L-shaped block with randomly sampled side lengths.

\begin{figure*}[!h]
            \centering
            \includegraphics[width=0.8\textwidth]{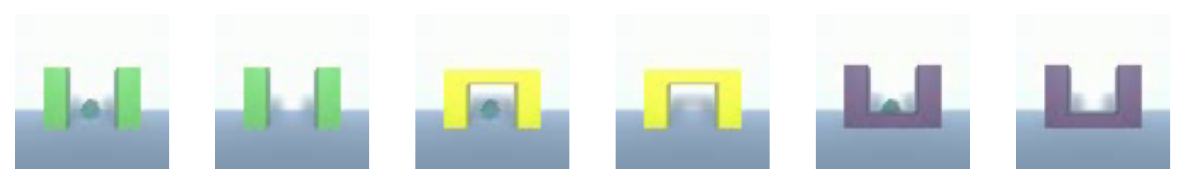}
            \vspace{-0.25cm}
            \caption{The middle frame for example sequences from the occlusion event data set. From left to right, the conditions are \emph{no occluder}, \emph{top occluder}, and \emph{bottom occluder}. Each condition consists of two sequences: the left sequence shows the expected sequence. The right sequence shows a violation where the moving object only appears on the outside of the occluders.}
            \label{fig:occlusion_test_example}
        \end{figure*}

    For the occlusion data set, the variations in the data set were: height of the pillars, height of the occluder, color of the occluder, and color of the moving object (see Figure \ref{fig:occlusion_test_example} for more exemplary test sequences). Additionally, in the training data set, the size of the moving object, the width of the pillars, the position of the occluder, and the speed of the moving object were varied. 

\begin{figure*}[!h]
            \centering
            \includegraphics[width=0.8\textwidth]{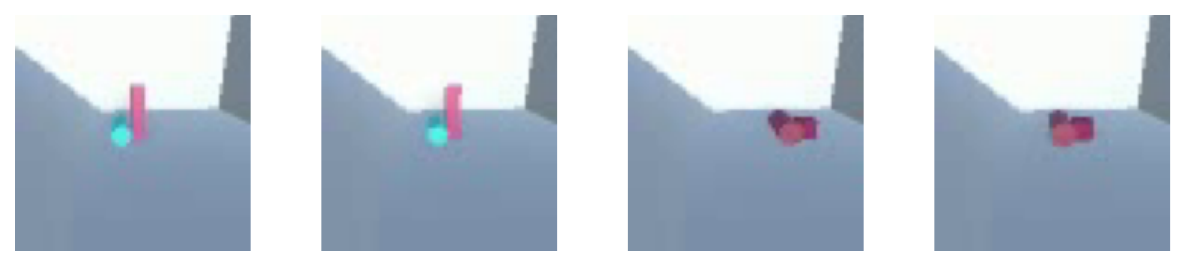}
            \caption{The last frame for example sequences from the collision event data set. From left to right, the conditions are \emph{displacement} and \emph{no vertical bias}. For each condition, there is a violated sequence left and an expected sequence right.}
            \label{fig:collision_test_example}
        \end{figure*}

    The variations in the collision event data sets were: stationary object size, moving object size, stationary object color, and moving object color (see Figure \ref{fig:collision_test_example} for more exemplary test sequences). In the training data set, the camera position and angle were also varied. 

    \clearpage
    \vspace*{0.1cm}
    \hspace{3.5cm}
    \textbf{Contact or no contact}
    \hspace{4.0cm}
    \textbf{Type of contact}

        \hspace{2.6cm}
        Violated
        \hspace{2.1cm}
        Expected
        \hspace{2.1cm}
        Violated
        \hspace{2.1cm}
        Expected

        \vspace{-0.3cm}
        \begin{figure*}[!h]
            \centering
            \includegraphics[width=0.8\textwidth]{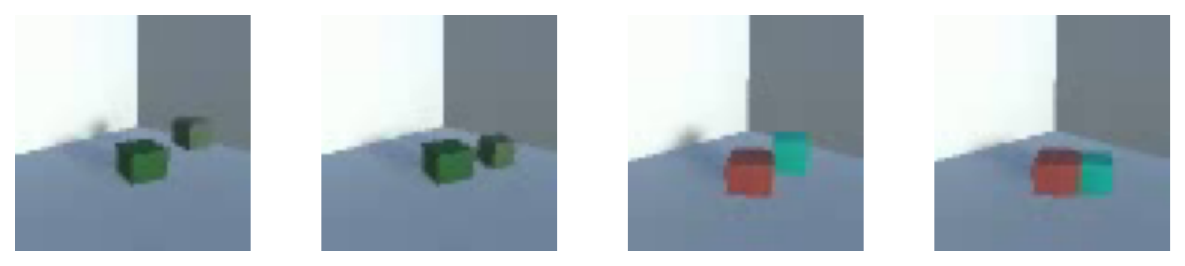}
        \end{figure*}

    \hspace{4.3cm}
    \textbf{Overlap}
    \hspace{5.8cm}
    \textbf{Shape}

        \hspace{2.6cm}
        Violated
        \hspace{2.1cm}
        Expected
        \hspace{2.1cm}
        Violated
        \hspace{2.1cm}
        Expected

        \vspace{-0.3cm}

    \vspace{0.6cm}
    \hspace{3.0cm}
    \textbf{No occluder}
    \hspace{2.5cm}
    \textbf{Top occluder}
    \hspace{2.2cm}
    \textbf{Bottom occluder}

        \hspace{2.0cm}
        Expected
        \hspace{0.9cm}
        Violated
        \hspace{0.9cm}
        Expected
        \hspace{0.9cm}
        Violated
        \hspace{0.9cm}
        Expected
        \hspace{0.9cm}
        Violated

        \vspace{-0.3cm}

    \vspace{0.6cm}
    \hspace{4.0cm}
    \textbf{Vertical bias}
    \hspace{4.9cm}
    \textbf{Displacement}

        \hspace{2.6cm}
        Violated
        \hspace{2.1cm}
        Expected
        \hspace{2.1cm}
        Violated
        \hspace{2.1cm}
        Expected

        \vspace{-0.3cm}

\clearpage
\section{Model reconstructions}
\label{app:recons}
    \subsection{Support events}
    \label{app:support_recons}

    \begin{figure*}[!h]
        \centering
        \includegraphics[width=1.0\textwidth]{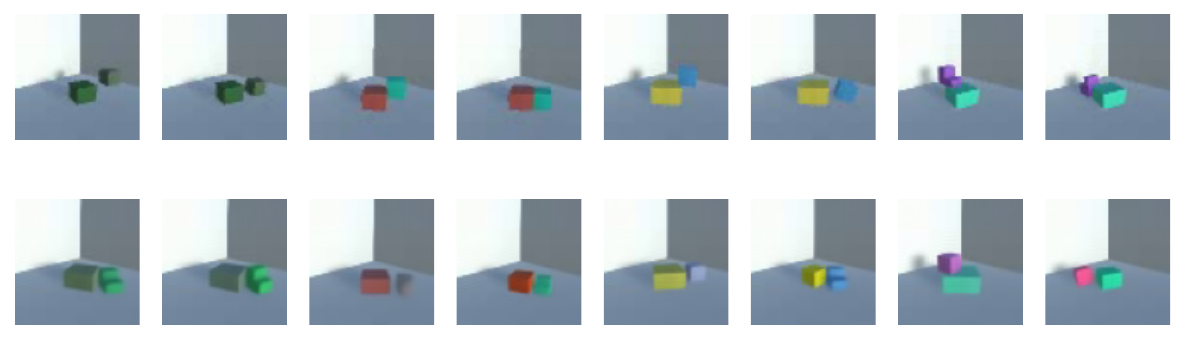}
        \caption{The top row shows the last frame from an example batch of the support event test data set. The bottom row shows the reconstructions by the model with $\beta = 1$ and using open loop reconstruction given only the first frame.}
        \label{fig:support_recon_gate1}
    \end{figure*}

    \begin{figure*}[!h]
        \centering
        \includegraphics[width=1.0\textwidth]{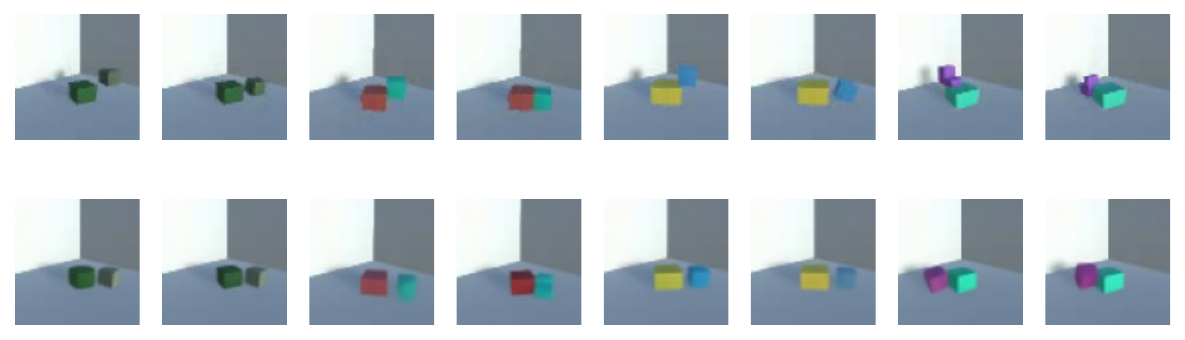}
        \caption{The top row shows the last frame from an example batch of the support event test data set. The bottom row shows the reconstructions by the model with $\beta = 1$ and using open loop reconstruction given the first two frames.}
        \label{fig:support_recon_gate2}
    \end{figure*}

    \begin{figure*}[!h]
        \centering
        \includegraphics[width=1.0\textwidth]{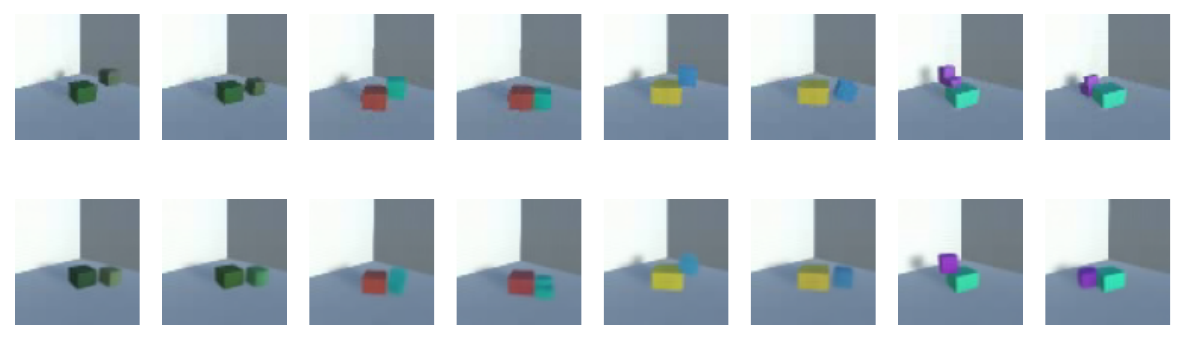}
        \caption{The top row shows the last frame from an example batch of the support event test data set. The bottom row shows the reconstructions by the model with $\beta = 1$ and using closed loop reconstruction.}
        \label{fig:support_recon_gate1000}
    \end{figure*}
    
    \clearpage
    \subsection{Occlusion events}
    \label{app:occlusion_recons}

    \vspace{-0.25cm}
    \begin{figure*}[!h]
        \centering
        \includegraphics[width=0.9\textwidth]{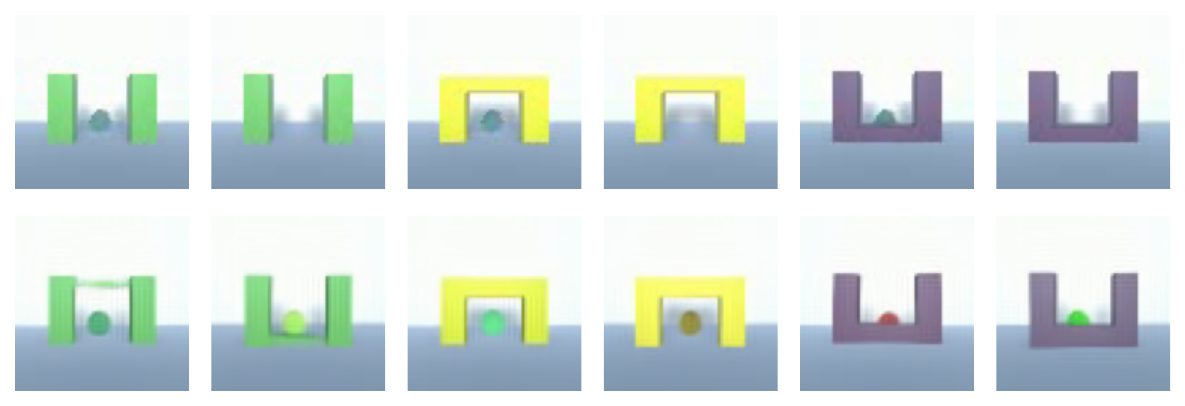}
        \vspace{-0.25cm}
        \caption{The top row shows the middle frame from an example batch of the occlusion event test data set. The bottom row shows the reconstructions by the model with $\beta = 1$ and using open loop reconstruction given only the first frame.}
        \label{fig:occlusion_recon_gate1}
    \end{figure*}
    
    \vspace{-0.25cm}
    \begin{figure*}[!h]
        \centering
        \includegraphics[width=0.9\textwidth]{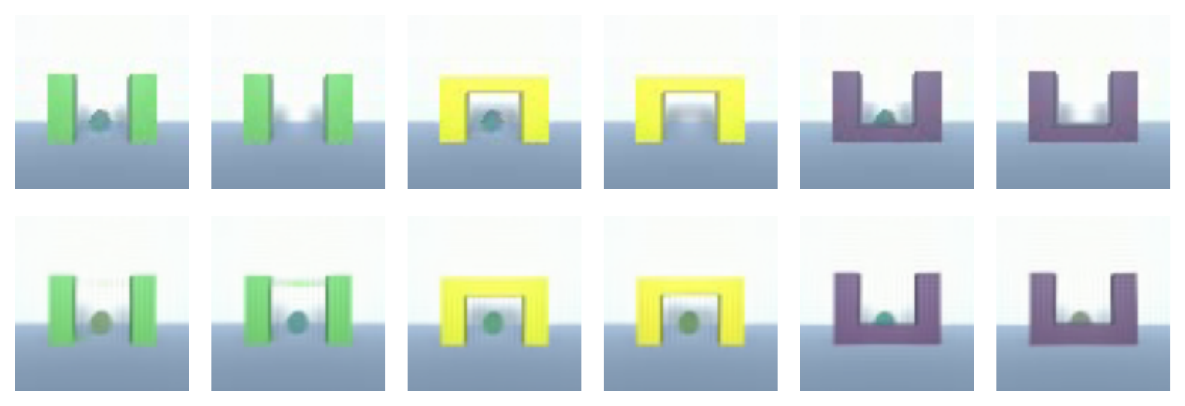}
        \vspace{-0.25cm}
        \caption{The top row shows the middle frame from an example batch of the occlusion event test data set. The bottom row shows the reconstructions by the model with $\beta = 1$ and using open loop reconstruction given the first two frames.}
        \label{fig:occlusion_recon_gate2}
    \end{figure*}

    \vspace{-0.25cm}
    \begin{figure*}[!h]
        \centering
        \includegraphics[width=0.9\textwidth]{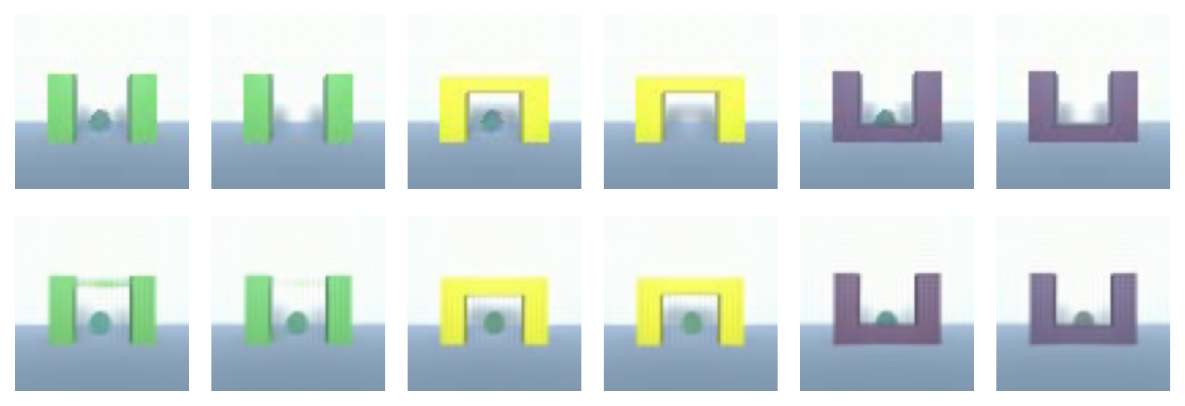}
        \vspace{-0.25cm}
        \caption{The top row shows the middle frame from an example batch of the occlusion event test data set. The bottom row shows the reconstructions by the model with $\beta = 1$ and using closed loop reconstruction.}
        \label{fig:occlusion_recon_gate1000}
    \end{figure*}
    
    \clearpage
    \subsection{Collision events}
    \label{app:collision_recons}
    
    \begin{figure*}[!h]
        \centering
        \includegraphics[width=1.0\textwidth]{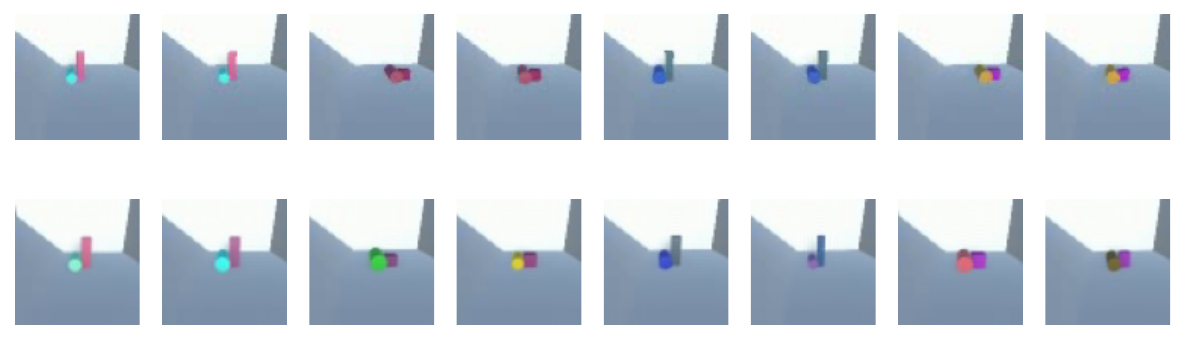}
        \caption{The top row shows the last frame from an example batch of the collision event test data set. The bottom row shows the reconstructions by the model with $\beta = 1$ and using open loop reconstruction given only the first frame.}
        \label{fig:collision_recon_gate1}
    \end{figure*}

    \begin{figure*}[!h]
        \centering
        \includegraphics[width=1.0\textwidth]{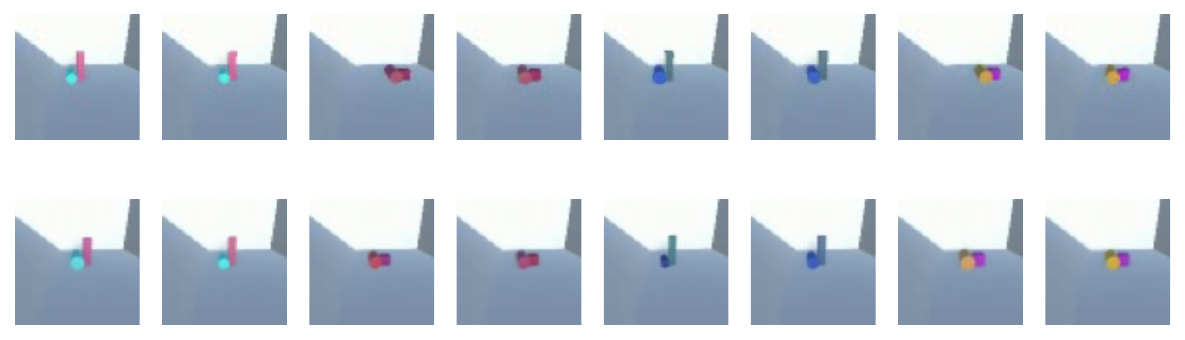}
        \caption{The top row shows the last frame from an example batch of the collision event test data set. The bottom row shows the reconstructions by the model with $\beta = 1$ and using open loop reconstruction given the first two frames.}
        \label{fig:collision_recon_gate2}
    \end{figure*}

    \begin{figure*}[!h]
        \centering
        \includegraphics[width=1.0\textwidth]{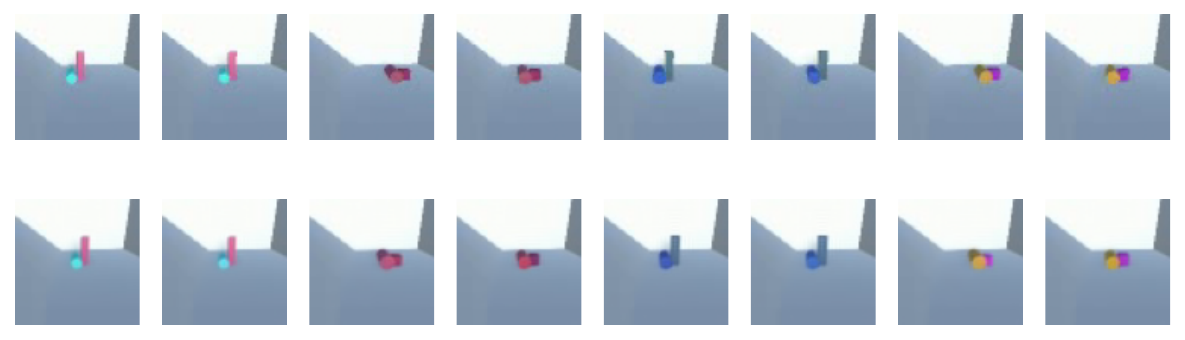}
        \caption{The top row shows the last frame from an example batch of the collision event test data set. The bottom row shows the reconstructions by the model with $\beta = 1$ and using closed loop reconstruction.}
        \label{fig:collision_recon_gate1000}
    \end{figure*}
    
\clearpage
\section{Potential negative societal impact}
\label{app:impact}
In the current work, we highlight an incongruency between the learning trajectories of current generative models and the developmental trajectories of children. As such, we do not expect there to be a potential negative societal impact as a consequence of this work. In contrast, we think that highlighting the differences between current machine learning algorithms and human cognition is important for a better understanding of the opportunities and dangers that may accompany artificial learning systems.

\section{Reconstruction errors}
\label{app:recon_errors}

\begin{figure*}[!h]
        \centering
        \includegraphics[width=1.0\textwidth]{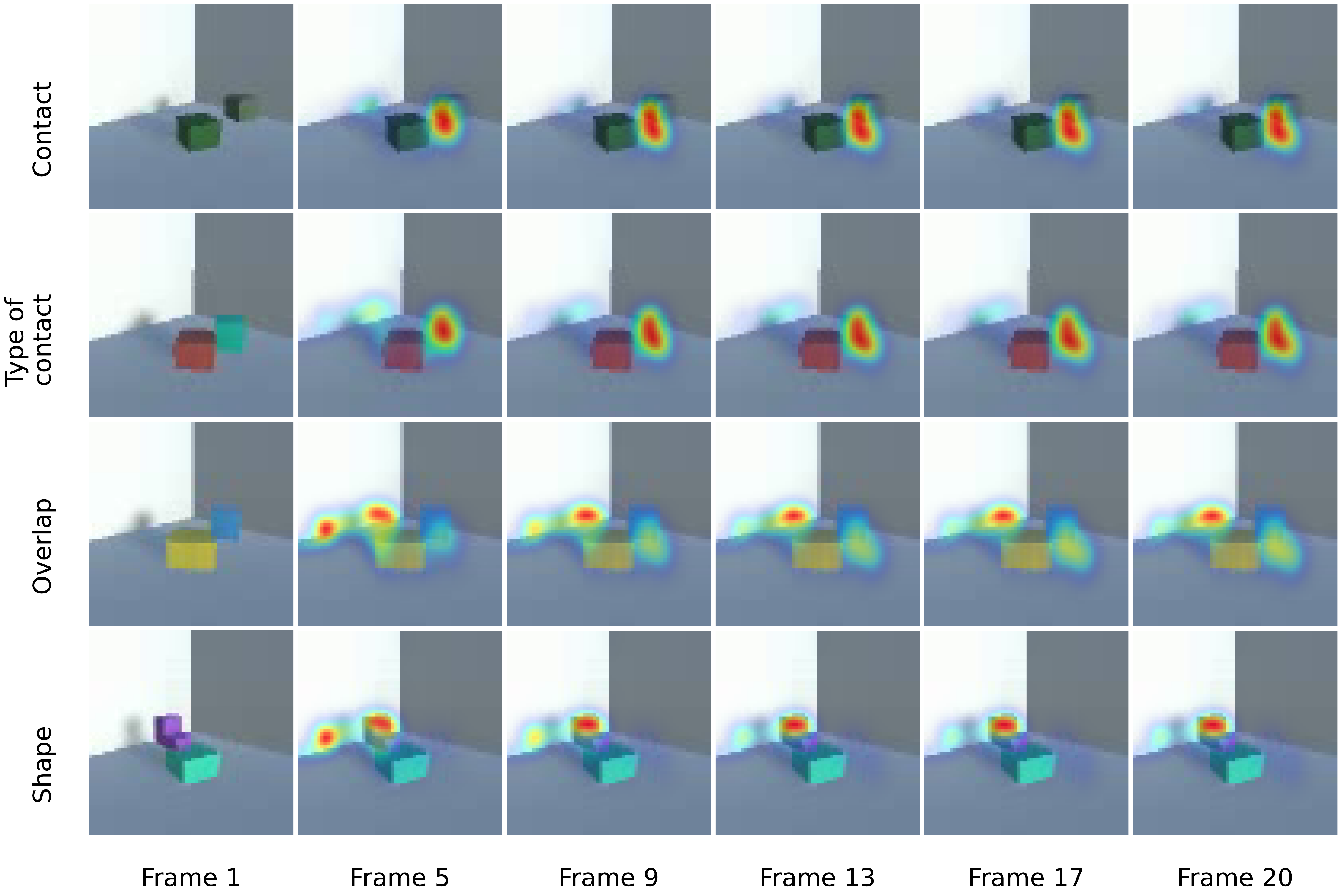}
        \caption{Frames from the violated sequences for one batch of the support event data sets with an overlay showing the negative log-likelihood of the observations under the model with $\beta = 1$ and using open loop reconstruction given the first two frames. We see that the model predominantly focuses on the actual and presumed location of the top cube as well as the shadow of the top cube.}
        \label{fig:support_recon_error}
    \end{figure*}

\begin{figure*}[!h]
        \centering
        \includegraphics[width=1.0\textwidth]{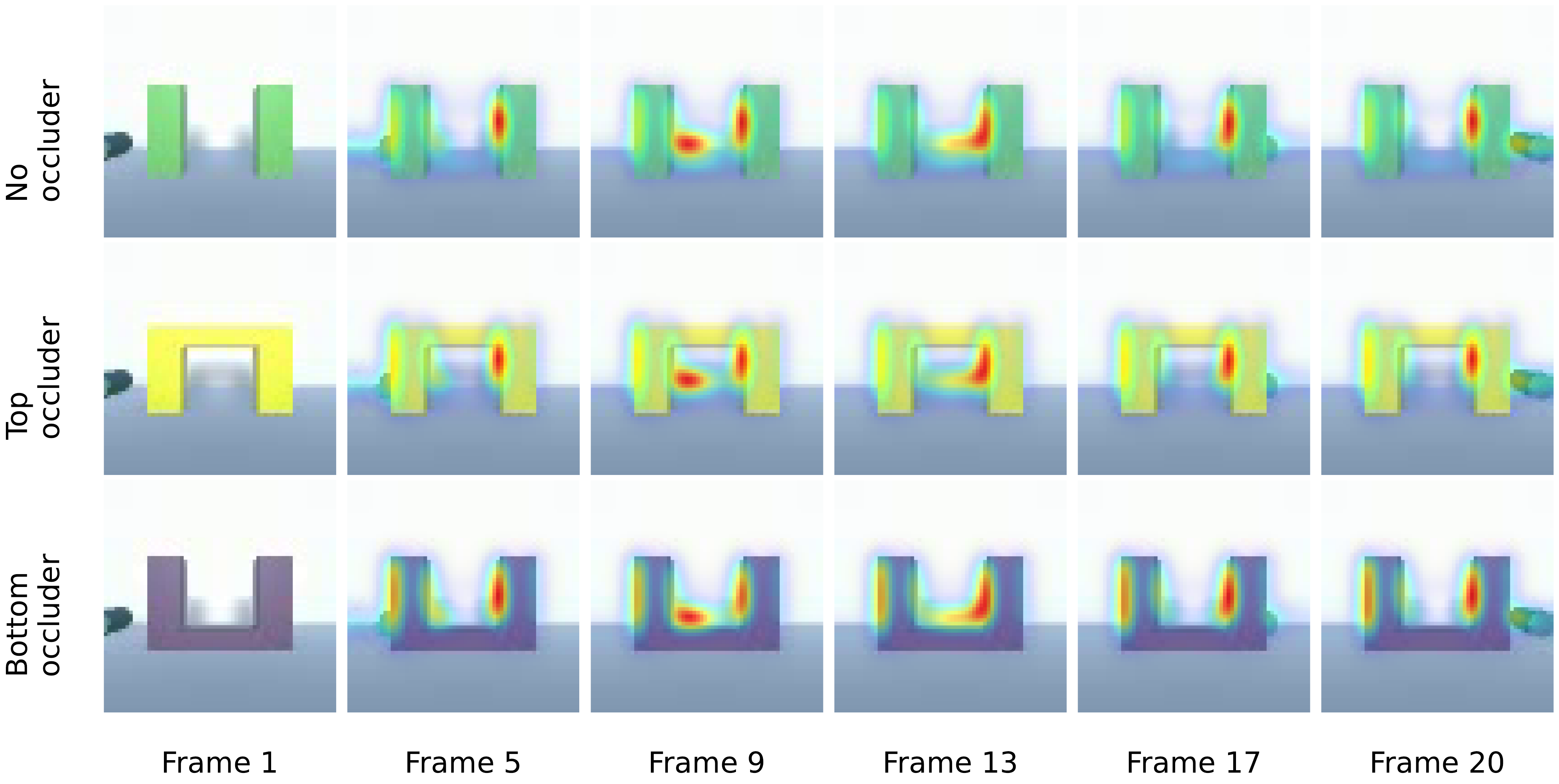}
        \caption{Frames from the violated sequences for one batch of the occlusion event data sets with an overlay showing the negative log-likelihood of the observations under the model with $\beta = 1$ and using open loop reconstruction given the first two frames. The violated sequences show a sphere moving behind an occluder and not reappearing in the gap between the columns. We see that the reconstruction error is large surrounding the edges of the columns. For the bottom occluder sequence, we also see an increased error on the edge of the connection between the two columns. }
        \label{fig:occlusion_recon_error}
    \end{figure*}

\begin{figure*}[!h]
        \centering
        \includegraphics[width=1.0\textwidth]{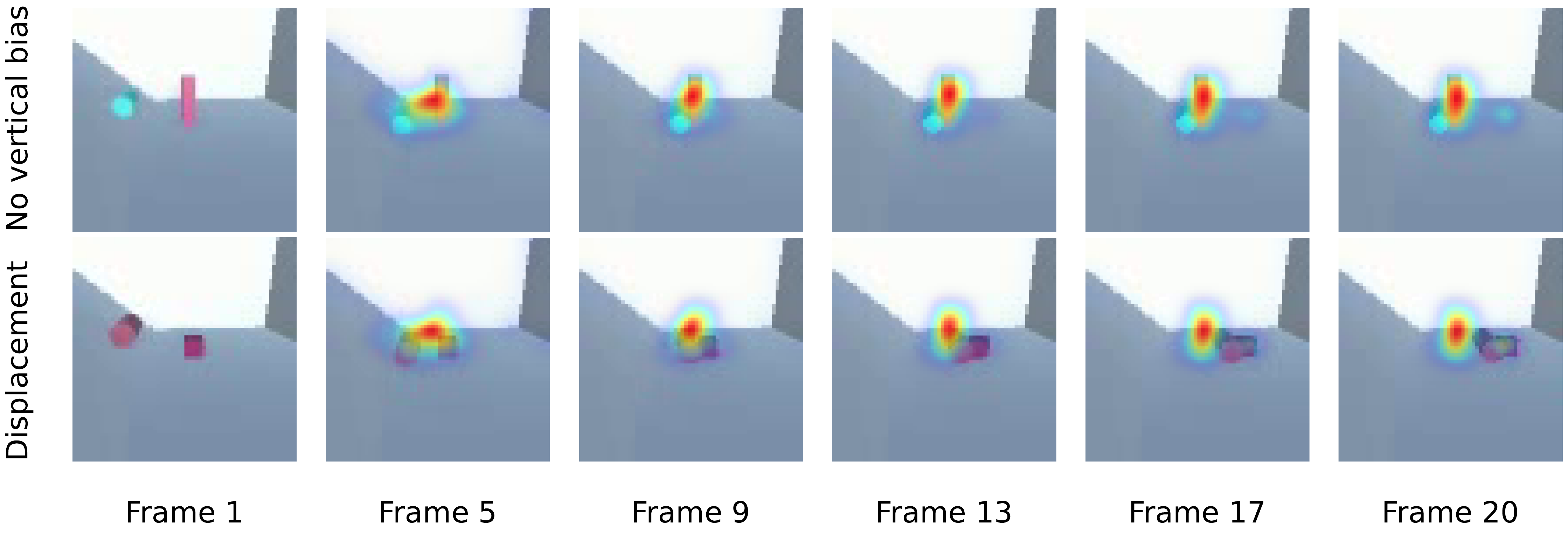}
        \caption{Frames from the violated sequences for one batch of the collision event data sets with an overlay showing the negative log-likelihood of the observations under the model with $\beta = 1$ and using open loop reconstruction given the first two frames. We see that the reconstruction error is especially large at the actual and presumed locations of the stationary object.}
        \label{fig:collision_recon_error}
    \end{figure*}

    Figures \ref{fig:support_recon_error}, \ref{fig:occlusion_recon_error}, and \ref{fig:collision_recon_error} show the reconstruction error for the respective event type data sets. The reconstruction error is given by the negative log-likelihood of the reconstructions given example violated sequences. It is displayed as an overlay on top of the violated sequences. 
    
    \subsection{Support events}
    \vspace{0.5cm}

    \clearpage
    \subsection{Occlusion events}

    \subsection{Collision events}

\clearpage
\section{Closed loop results}
\label{app:closed}

    \vspace{-0.3cm}
     \begin{figure*}[!h]
        \centering
        \includegraphics[width=1.0\textwidth]{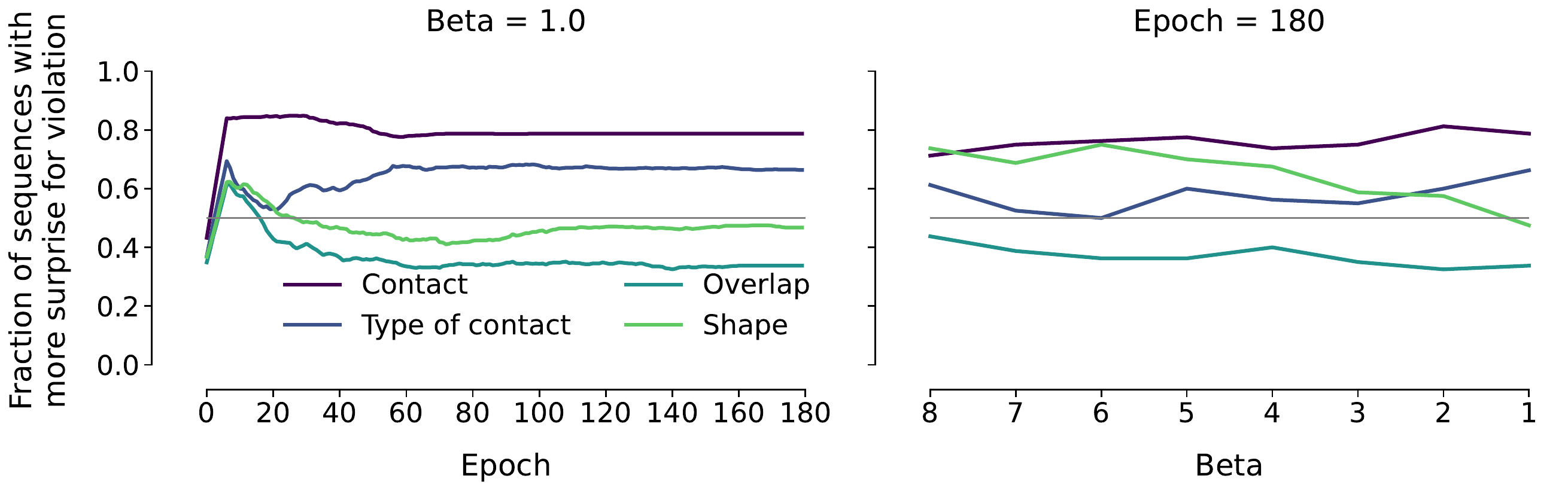}
        \vspace{-0.9cm}
        \caption{This is the closed loop counterpart to Figure \ref{fig:support_hypotheses}. The plot on the left shows the percentage of sequences for which the surprise for the violated sequence exceeds that of the expected sequence for the model with $\beta = 1$ at every epoch and for each condition of the support event data set separately. The lines are smoothed with a uniform kernel of size 10. The plot on the right shows the same metric for fully trained models with different $\beta$.}
        \label{fig:support_main_closed}
    \end{figure*}

    \vspace{-0.3cm}
    \begin{figure*}[!h]
        \centering
        \includegraphics[width=1.0\textwidth]{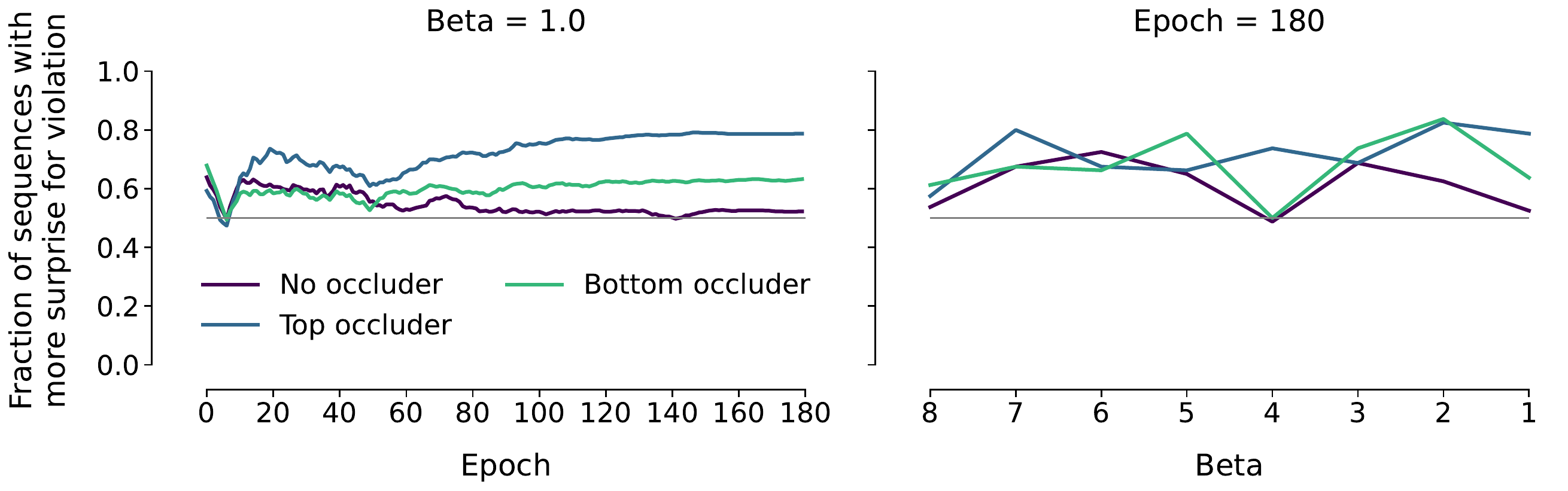}
        \vspace{-0.9cm}
        \caption{This is the closed loop counterpart to Figure \ref{fig:occlusion_hypotheses}. The plot on the left shows the percentage of sequences for which the surprise for the violated sequence exceeds that of the expected sequence for the model with $\beta = 1$ at every epoch and for each condition of the occlusion event data set separately. The lines are smoothed with a uniform kernel of size 10. The plot on the right shows the same metric for fully trained models with different $\beta$.}
        \label{fig:occlusion_main_closed}
    \end{figure*}

    \vspace{-0.3cm}
    \begin{figure*}[!h]
        \centering
        \includegraphics[width=1.0\textwidth]{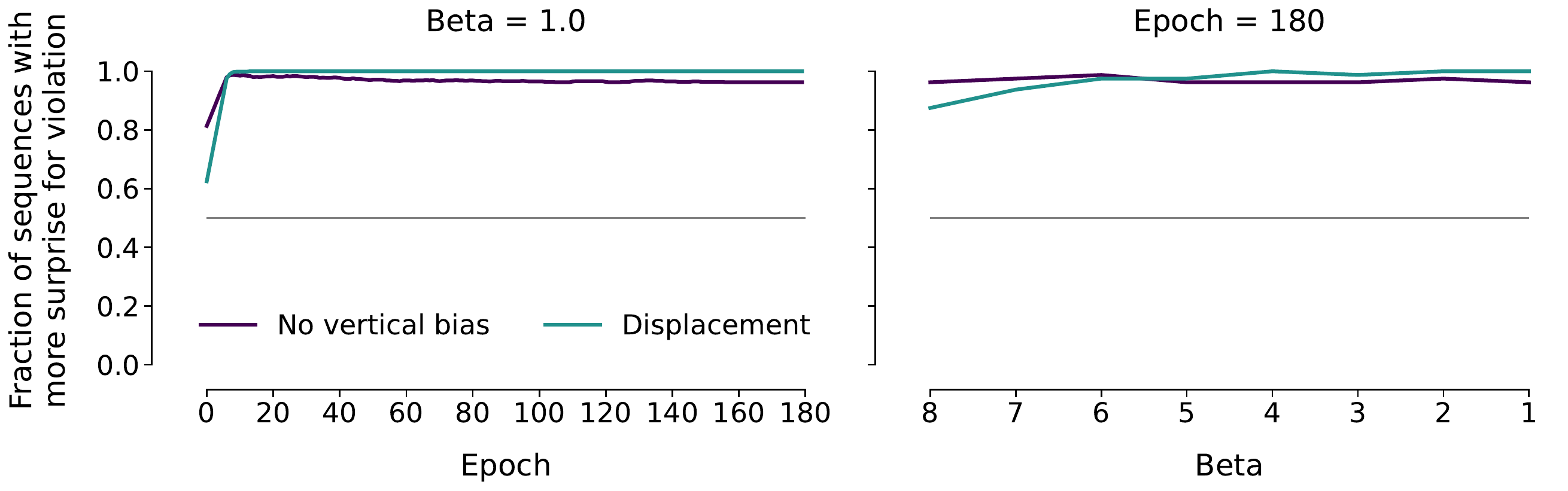}
        \vspace{-0.9cm} 
        \caption{This is the closed loop counterpart to Figure \ref{fig:collision_hypotheses}. The plot on the left shows the percentage of sequences for which the surprise for the violated sequence exceeds that of the expected sequence for the model with $\beta = 1$ at every epoch and for each condition of the collision event data set separately. The lines are smoothed with a uniform kernel of size 10. The plot on the right shows the same metric for fully trained models with different $\beta$.}
        \label{fig:collision_main_closed}
    \end{figure*}

\clearpage
\section{Minimal model-based hidden layer analysis}
\label{app:tsne}

    \begin{figure*}[!h]
        \centering
        \includegraphics[width=0.6\textwidth]{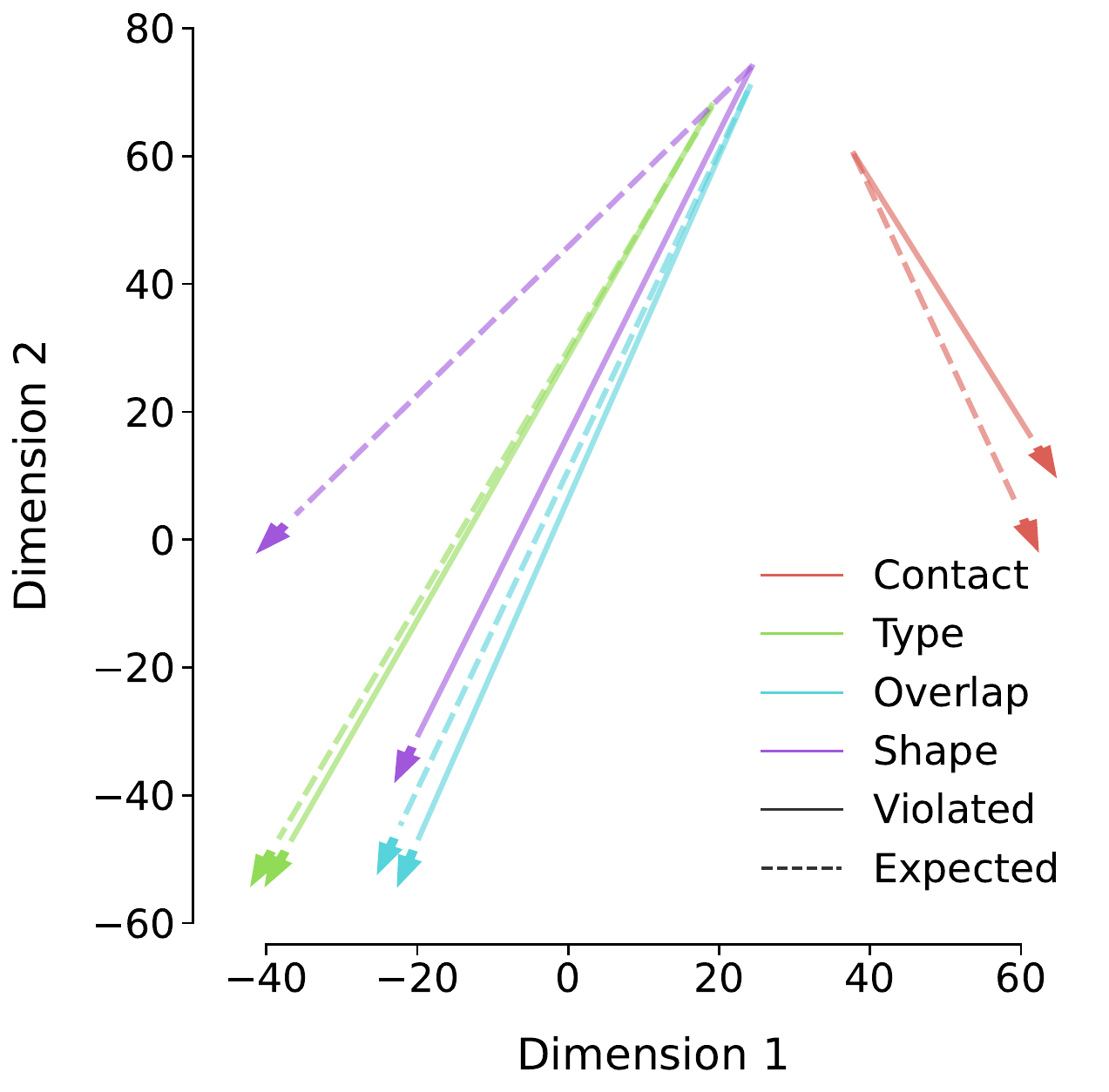}
        \caption{This plot shows the lower dimensional embedding of the hidden layer activations for the first and last frame of example test sequences in the support data set.}
        \label{fig:tsne}
    \end{figure*}
    
Based on the recommendation of a reviewer, we decided to perform a minimal model-based analysis of the hidden layers. We here report a t-SNE analysis similar to the one performed in Appendix B of \citet{piloto2018probing}. We first created a lower dimensional embedding of the hidden layer activations for the frames of the sequences in the support test data set. We then plotted the embeddings of the first and last frames of an exemplary violated and expected sequence for each of the four conditions in the support test data set (see Figure \ref{fig:tsne}).

For conditions in which the surprise for the violated sequences exceeded that for the expected sequences for a majority of the test data set, such as the "contact" and "shape" conditions, we see that the latent representations for the violated and the expected sequence diverge significantly in the later frames of the example sequences. In contrast, the separation is weaker for conditions where the model performs less well. This points towards the model not having learned sufficiently separated representations in order to properly distinguish physically implausible and plausible sequences.

We want to highlight that this analysis is rather minimal and rudimentary. Future work should ideally analyze the latent representations of generative models over time in order to better understand why they not only fail to learn specific physical concepts but also why their learning diverges from that of children.

\end{document}